\pdfoutput=1
\pdfoutput=1
\pdfoutput=1
\pdfoutput=1
\pdfoutput=1

\documentclass{article}
\usepackage{ijcai24}

\usepackage{times}
\usepackage{soul}
\usepackage{url}
\usepackage[hidelinks]{hyperref}
\usepackage[utf8]{inputenc}
\usepackage[small]{caption}
\usepackage{graphicx}
\usepackage{amsmath}
\usepackage{amsthm}
\usepackage{booktabs}
\usepackage{algorithm}
\usepackage{algorithmic}
\usepackage[switch]{lineno}

\usepackage{amsmath}
\usepackage{amssymb}
\usepackage{graphicx}  
\usepackage{subcaption}
\usepackage{booktabs}
\usepackage{multirow}

\usepackage{changes}

\newtheorem{theorem}{Theorem}
\newtheorem{lemma}{Lemma}
\newtheorem{corollary}{Corollary}
\newtheorem{proposition}{Proposition}
\newtheorem{properties}{Properties}

\title{Exterior Penalty Policy Optimization with Penalty Metric Network under Constraints}

\author{
Shiqing Gao
\and
Jiaxin Ding\footnote{Corresponding Author.}\and
Luoyi Fu\and
Xinbing Wang\And
Chenghu Zhou
\affiliations
Shanghai Jiao Tong University\\
\emails
\{shiqinggao, jiaxinding, yiluofu, xwang8\}@sjtu.edu.cn,
zhouchsjtu@gmail.com
}

\begin{document}

\maketitle

\begin{abstract}
    In Constrained Reinforcement Learning (CRL), agents explore the environment to learn the optimal policy while satisfying constraints. The penalty function method has recently been studied as an effective approach for handling constraints, which imposes constraints penalties on the objective to transform the constrained problem into an unconstrained one. However, it is challenging to choose appropriate penalties that balance policy performance and constraint satisfaction efficiently. In this paper, we propose a theoretically guaranteed penalty function method, Exterior Penalty Policy Optimization (EPO), with adaptive penalties generated by a Penalty Metric Network (PMN). PMN responds appropriately to varying degrees of constraint violations, enabling efficient constraint satisfaction and safe exploration. We theoretically prove that EPO consistently improves constraint satisfaction with a convergence guarantee. We propose a new surrogate function and provide worst-case constraint violation and approximation error. In practice, we propose an effective smooth penalty function, which can be easily implemented with a first-order optimizer. Extensive experiments are conducted, showing that EPO outperforms the baselines in terms of policy performance and constraint satisfaction with a stable training process, particularly on complex tasks.
\end{abstract}

\section{Introduction}

Deep reinforcement learning (RL) \cite{sutton2018reinforcement} has demonstrated great potential in various real-world scenarios, including video games \cite{mnih2016asynchronous,wang2016sample,vinyals2019grandmaster}, robotics control \cite{gu2016continuous,levine2016end,haarnoja2018soft}, and Go \cite{silver2016mastering,silver2017mastering}. 
However, many real-world problems inherit constraints when finding the optimal policy. For example, a robot agent should avoid obstacles while completing tasks. Such policy optimization with constraints is the focus of constrained RL (CRL), a critical field of RL research that expands the applicability of RL in real-world scenarios.

CRL is generally modeled as Constrained Markov Decision Processes (CMDP) \cite{beutler1985optimal,ross1989markov,altman2021constrained}, in which agents aim to maximize rewards while satisfying constraints. CMDP integrates safety criteria in the form of constraints into RL and provides a fundamental mathematical framework.

Extensive research has been conducted in CMDP, but most existing methods have inherent oscillations or high computational complexity.
CPO \cite{achiam2017constrained} approximates the objective with a second-order expansion, leading to high computation costs of the high-dimensional inverse Fisher information matrix. 
Primal-dual methods convert constrained problems into unconstrained ones using the Lagrangian function to solve in the dual space. However, the learning dynamics exhibit high oscillations and constraint violations \cite{tessler2018reward,yu2019convergent,paternain2019constrained,ding2020natural}.
APPO \cite{dai2023augmented} employs the augmented Lagrangian method to enhance the stability of the primal-dual algorithm, but it is sensitive to Lagrangian multiplier initialization \cite{chow2018lyapunov,stooke2020responsive}.
Projection-based methods, such as PCPO \cite{yang2020projection} and FOCOPS \cite{zhang2020first}, project actions back to the feasible region, presenting a costly computation and underperforming cumulative rewards \cite{zhang2023evaluating}.

Penalty function methods impose constraints as penalties on the objective, addressing CRL in the primal space without introducing dual variables or requiring a complex second-order Taylor expansion \cite{liu2020ipo}. However, it is challenging to choose an appropriate penalty to trade off superior performance and efficient constraint satisfaction. An excessive penalty leads to a conservative policy that fails to explore the environment, while a minor penalty cannot guarantee the safety of the policy \cite{zhang2023evaluating}.
IPO \cite{liu2020ipo} restricts the policy within feasible regions with a logarithmic barrier function, which outputs excessive penalties near boundaries of the feasible region and cannot handle tasks where the initial policy violates constraints. P3O \cite{zhang2022penalized} utilizes a linear penalty function with the ReLU operator to optimize policy. When constraints are severely violated, the linear relation cannot provide sufficient penalty to bring the policy back to the feasible region quickly.

In this paper, we divide the policy space based on the degree of constraint violation and introduce a Penalty Metric Network (PMN) comprising linear and quadratic cost value critics to capture penalties in different policy regions.
The linear critic provides consistent incremental responses to small violations, while the quadratic critic imposes more severe penalties for larger violations, strongly discouraging substantial breaches.
A dedicated weighting layer within PMN is used to aggregate different types of penalties, producing suitable penalties for efficient constraint satisfaction. 
A filter layer eliminates penalties in the feasible region, ensuring safe exploration for optimal performance.
We propose a first-order penalty function algorithm, Exterior Penalty Policy Optimization (EPO), which imposes penalties output from PMN on the objective to transform constrained policy optimization into an unconstrained one. 
EPO guarantees monotonically non-increasing constraints through convergence analysis by maximizing sequential subproblems. 
Furthermore, we present a surrogate penalty function using samples generated from the current policy and theoretically provide the worst-case constraint violation and approximation error.
In practice, EPO employs a smooth surrogate penalty function and an adaptive penalty factor adjustment strategy to ensure a stable training process.
Extensive experiments on challenging tasks demonstrate that EPO achieves faster constraint satisfaction and better policy performance compared to multiple baseline algorithms.

In summary, our contributions are as follows:
\begin{itemize}
    \item We design the Penalty Metric Network (PMN) to capture suitable constraint penalties in different policy regions. 
    \item We propose the EPO algorithm based on PMN, which guarantees stable convergence and efficient constraint satisfaction. Theoretical guarantees on worst-case constraint violation and approximation errors are provided.
    \item We conduct extensive experiments to compare EPO with multiple state-of-the-art algorithms. EPO outperforms the baselines with higher performance and better constraint satisfaction.
\end{itemize}

\section{Related Work}

In the tabular case, the CMDP problem has been extensively studied \cite{beutler1986time,ross1989randomized,ross1991multichain}. For high-dimensional problems, RL with constraints is a challenging topic. Many methods have been proposed in recent years, which can be roughly divided into primal methods and primal-dual methods.

\paragraph{Primal-Dual methods.} Primal-dual methods transform constrained problems into unconstrained ones by introducing dual variables \cite{chow2017risk,ding2021provably}. \cite{tessler2018reward} propose multi-timescale Lagrangian methods, which utilize penalty signals to guide the policy update towards constraint satisfaction. The traditional Lagrangian multiplier update behaves as integral control, and \cite{stooke2020responsive} introduce proportional and differential control to reduce cost overshoot and oscillations in the learning dynamics. \cite{ding2020natural} establish the global convergence with sublinear rates regarding the optimality gap.
The Augmented Lagrangian method is introduced in \cite{dai2023augmented} to reduce the oscillations during training.
However, primal-dual approaches are still sensitive to initial parameters, limiting their applicability \cite{zhang2022penalized}.

\paragraph{Primal methods.} Primal methods directly optimize the constrained problem in the primal space \cite{chow2018lyapunov,chow2019lyapunov,dalal2018finite,yu2022towards}. CPO \cite{achiam2017constrained} provides a lower bound on performance and an upper bound on constraint violation. PCPO \cite{yang2020projection} first improves the reward within the trust region, then projects the policy to the feasible region. However, second-order methods in CPO and PCPO, which are based on local policy search \cite{peters2008reinforcement}, encounter computational challenges due to the high-dimensional Hessian matrix. FOCOPS \cite{zhang2020first} proposes a first-order policy optimization method, which first solves the constraint problem in the nonparametric policy space and then projects the updated policy back into the parametric space. CUP \cite{yang2022constrained} provides generalized theoretical guarantees for surrogate functions with a generalized advantage estimator \cite{schulman2015high}, effectively reducing variance while maintaining acceptable bias. However, projection approaches have poor performance since the correction only ensures feasibility but lacks equivalence with optimality \cite{zhang2023evaluating}.
All the above methods involve solving primal-dual subproblems, and the introduced dual variables lead to an unstable training process. Our approach offers a simpler first-order optimization method in the primal space, ensuring the stability of constraint satisfaction.

IPO \cite{liu2020ipo} augments the objective with a logarithmic barrier function to restrict the policy to feasible regions. However, the logarithmic function becomes ineffective when constraints are violated, limiting the agent's exploration. P3O \cite{zhang2022penalized} penalizes constraints with a ReLU operator to obtain an unconstrained problem. However, the linear penalty function in P3O cannot provide significant penalties for severe constraint violations, and the ReLU operator is non-differentiable at constraint boundaries, resulting in inaccurate gradient information.
Compared to P3O and IPO, our PMN provides more appropriate penalties based on policy regions for faster constraint satisfaction. Theoretically, EPO provides convergence guarantees, continuous improvement constraints, and approximation errors, guaranteeing better performance and constraint satisfaction.

\section{Preliminaries}

RL can be modeled as a Markov decision process (MDP), represented by a tuple $(S,A,R,\mathcal{P},\rho,\gamma)$, where $S$ denotes the state space, $A$ denotes the action space, $R:S \times A \rightarrow \mathbb{R}$ is the reward function, $\mathcal{P}:S \times A \rightarrow [0,1] $ is the transition probability function, $\rho$ is the initial state distribution, and $\gamma \in (0,1)$ is the discount factor of the reward. The policy $\pi: S \rightarrow A$ maps states to probability distributions over actions, where $\pi(a|s)$ is the probability of choosing action $a$ in state $s$, and $\Pi$ is the set of all stationary policies. The discounted future state visitation distribution is defined as $d^\pi(s):=(1-\gamma)\sum_{t=0}^\infty \gamma^t \mathcal{P}(s_t=s|\pi)$. 
In RL, we aim to find the optimal policy by maximizing the expected discounted return $J_R(\pi):=\mathbb{E}_{\tau \sim \pi} [\sum_{t=0}^{\infty} \gamma^t R(s_t,a_t)] $, where $\tau=(s_0,a_0,s_1,a_1,\cdots)$ denotes the trajectory based on policy $\pi$. The value function based on policy $\pi$ is $V_R^{\pi}(s):=\mathbb{E}_{\tau \sim \pi} [\sum_{t=0}^{\infty} \gamma^t R(s_t, a_t)|s_0=s]$, and action-value function is $Q_R^{\pi}(s, a):=\mathbb{E}_{\tau \sim \pi} [\sum_{t=0}^{\infty} \gamma^t R(s_t, a_t)|s_0=s,a_0=a]$. The advantage function measures the advantage of action $a$ over the mean value: $A_R^{\pi}(s, a):=Q_R^\pi(s,a)-V_R^\pi(s)$.

The CMDP $(S,A,R,C,\mathcal{P},\rho,\gamma)$ introduces constraints to the MDP to restrict allowable policies. $C: S \times A \rightarrow \mathbb{R}$ is the cost function, mapping state-action pairs to costs. $d$ denotes the constraints threshold, the expected cumulative discount cost aims to satisfy $J_C(\pi):=\mathbb{E}_{\tau \sim \pi} [\sum_{t=0}^{\infty} \gamma^t C(s_t,a_t)] \leq d$. The cost value function $V_C^\pi(s)$, cost action-value function $Q_C^\pi(s, a)$ and cost advantage function $A_C^\pi(s, a)$ in CMDP can be obtained as in MDP by replacing the reward $R$ with the cost $C$. The CRL aims to find an optimal policy by maximizing the expected discount return over the set of feasible policies $\Pi_C:=\{\pi\in\Pi\ :J_C(\pi)\leq d\}$:

\begin{equation}
\label{eq1}
\begin{aligned}
    \arg\max\limits_{\pi\in\Pi} &J_R(\pi) \\
    s.t. \quad & J_C(\pi) \leq d
\end{aligned}
\end{equation}

The following equation gives the performance difference between two arbitrary policies \cite{kakade2002approximately}:

\begin{equation}
\label{eq2}
    J_R(\pi')-J_R(\pi)=\frac{1}{1-\gamma}\mathbb{E}_{{s\sim{d^{\pi'}}},a\sim{\pi'}}[A_R^{\pi}(s,a)]
\end{equation}
This implies that iterative updates to the policy, $\pi'(s)=\arg\max_a A_R^\pi(s,a)$, lead to performance improvement until convergence to the optimal solution.

\section{Methodology}

In this section, we present the EPO algorithm and its theoretical foundation. 
We first show the structure of the PMN. Then, we propose the EPO method based on the PMN. The convergence analysis and error bound are provided to ensure effectiveness. Finally, we propose a smooth penalty function with penalty factor terms, which can be easily optimized in practice using a first-order optimizer.

\begin{figure}[h]
    \centering
         \includegraphics[width=0.47\textwidth]{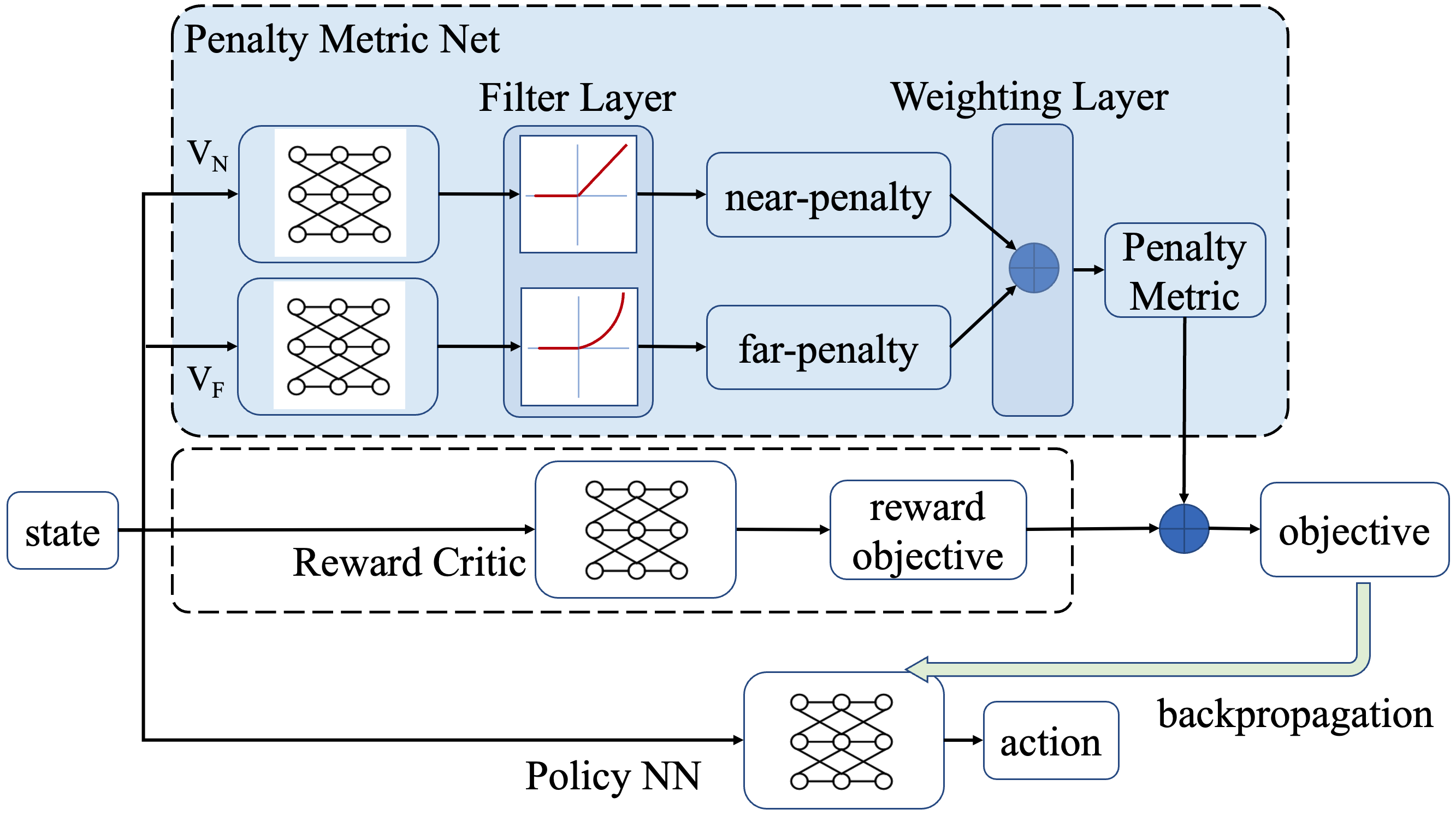}
     \hfill
    \caption{Structure of the Penalty Metric Network in EPO, comprising two streams to capture the near-penalty and far-penalty. $V_N$ and $V_F$ denote the linear and quadratic cost value critics, respectively. The Weighting Layer integrates the different types of penalties and outputs the Penalty Metric, which is imposed on the objective to guide the policy update.}
    \label{fig:structure}
\end{figure}

\subsection{Exterior Penalty Function with Penalty Metric Network}
According to the performance difference equation (\ref{eq2}), our problem is defined as a constrained optimization problem:

\begin{equation}
\label{eq6}
\begin{aligned}
    \pi_{k+1}= & \arg\max \limits_{\pi \in {\Pi_\theta}} \mathbb{E}_{s\sim{d^\pi},a\sim\pi} [A_R^{\pi_k}(s,a)]  \\
    s.t. \quad & J_C(\pi_k)+\frac{1}{1-\gamma} \mathbb{E}_{s\sim{d^\pi},a\sim\pi} [A_C^{\pi_k}(s,a)] \leq d \\
\end{aligned}
\end{equation}
where policy $\pi \in {\Pi_\theta}$ is parameterized with parameters $\theta$, and $\pi_k$ represents the current policy.
For simplification, we introduce the notations to represent the above objective function $F_R(\pi) := \mathbb{E}_{s\sim{d^\pi},a\sim\pi} [A_R^{\pi_k}(s,a)]$ and constraint function $F_C(\pi) := J_C(\pi_k)+\frac{1}{1-\gamma} \mathbb{E}_{s\sim{d^\pi},a\sim\pi} [A_C^{\pi_k}(s,a)] - d$.

$F_C(\pi)$ can be considered as a measure of the constraint distance between policy $\pi$ and the constraint boundaries in the policy space, indicating the degree of constraint violation.
We normalize $F_C(\pi)$ across samples in the current replay buffer for uniform scaling. Then we divide the policy space into near-boundary region and far-boundary region according to $F_C(\pi)$, defining the penalties within these regions as near-penalty and far-penalty, respectively.

We propose a Penalty Metric Network to capture penalties across various regions.
The PMN consists of two streams, representing the near-penalty and the far-penalty, as shown in Figure \ref{fig:structure}. 
We construct the linear cost value critic $V_N(s)$ and the quadratic cost value critic $V_F(s)$ as the backbone of near-penalty and far-penalty streams, respectively. 
Both critics are designed to provide sufficient penalties in their respective regions, ensuring efficient constraint satisfaction. 
For sufficient safety exploration, a filter layer eliminates negative values in both critics to ensure the penalty in the feasible region is 0. 
A weighting layer then integrates the two streams to generate a penalty metric for the current policy. We denote the PMN as a function $\Phi(F_C^+(\pi))$:

\begin{equation}
\label{eq_pmn}
    \Phi(F_C^+(\pi)):=\alpha F_C^+(\pi) + (1-\alpha)(F_C^+(\pi))^2
\end{equation}
where $\alpha\in[0,1]$ is the weight factor in weighting layer, and $F_C^+(\pi) = \max\{(F_C(\pi)), 0\}$, $\max\{\cdot\}$ represents the filter operation.
Imposing the penalty metric $\Phi(F_C^+(\pi))$ to the objective function $F_R(\pi)$, we obtain the exterior penalty function $P(\pi, \mu)$, which transforms the original constrained problem into an unconstrained one:

\begin{equation}
\label{eq10}
\begin{aligned}
    P(\pi,\mu)=&F_R(\pi)-\frac{1}{\mu}\Phi(F_C^+(\pi))\\
\end{aligned}
\end{equation}
where $\mu>0$ denotes the penalty factor. 
Now we get the new objective for (\ref{eq6}) by maximizing the penalty function:

\begin{equation}
\label{eq9}
\begin{aligned}
    \arg\max \limits_{\pi \in {\Pi_\theta}} & F_R(\pi)-\frac{1}{\mu}\left(\alpha F_C^+(\pi) + (1-\alpha)(F_C^+(\pi))^2\right) \\
\end{aligned}
\end{equation}

When constraints are satisfied, the objective focuses solely on maximizing the expected return. Conversely, when constraints are violated, the objective shifts to maximizing the expected return while simultaneously minimizing the cost.

Our Penalty Metric Network with linear and quadratic value estimators can provide appropriate penalties and bring the policy back to the feasible region $\Pi_C$ quickly. 

\begin{properties}
\label{properties1}
    Normalize the $F_C(\pi)$ and define the policy space for $F_C(\pi) \le \frac{1}{2}$ as the near-boundary region, for $F_C(\pi) > 1$ as the far-boundary region, and for $ \frac{1}{2}< F_C(\pi) \le 1$ as the mid-boundary region. 
    Increasing $\alpha$ provides increased penalty $\Phi(F_C^+(\pi))$ and gradient $|\nabla_\pi \Phi(F_C^+(\pi))|$ in the near-boundary region. Decreasing $\alpha$ provides increased penalty $\Phi(F_C^+(\pi))$ and gradient $|\nabla_\pi \Phi(F_C^+(\pi))|$ in the far-boundary region.
    Increasing $\alpha$ provides increased penalty $\Phi(F_C^+(\pi))$ and reduced gradient $|\nabla_\pi \Phi(F_C^+(\pi))|$ in the mid-boundary region.
\end{properties}

The weight $\alpha$ in the weighting layer is chosen based on $F_C(\pi)$.
This ensures that PMN consistently generates large penalties and gradients when constraints are violated, enabling the policy to quickly return to the feasible region.
In this paper, we set $\alpha=0$ in the far-boundary region, 1 in the near-boundary region, and 0.5 in the mid-boundary region.

We use two neural networks $V_N(\upsilon)$ and $V_{F}(\psi)$, parameterized by $\upsilon$ and $\psi$ respectively, to capture the near-boundary cost value and far-boundary cost value.
The Penalty Metric Network is updated by:

\begin{equation}
\label{eqCostCcritic}
\resizebox{0.47\textwidth}{!}{$
\begin{aligned}
    &\arg\min \limits_{\upsilon,\psi} \mathbb{E}_{s\sim{d^\pi},a\sim\pi} \left[ (V_N^{target} - V_N(s_t;\upsilon))^2\right]\\
    &+ \mathbb{E}_{s\sim{d^\pi},a\sim\pi} \left[ (V_{F}^{target} + V_{F}(s_t;\psi) - 2\sqrt{V_{F}^{target}V_{F}(s_t;\psi)}) \right]
\end{aligned}
$}
\end{equation}
where $V_N^{target} = \max\{C_t + \gamma V(s_{t+1}) - d, 0 \}$ and $V_F^{target} = (\max\{C_t + \gamma V(s_{t+1}) -d, 0 \})^2$ are targets for $V_N$ and $V_F$.

\subsection{Exterior Penalty Policy Optimization and Convergence Analysis}
Consider a decreasing sequence of penalty factors $\{\mu_t\}$ with $\mu_t\rightarrow 0$ as $t\rightarrow \infty$, where $t$ denotes the iteration number of penalty factor. For each penalty factor $\mu_t$, we solve a maximization problem of the penalty function $P(\pi, \mu_t)$ to obtain the corresponding global maximum policy $\pi_t$. 
By decreasing the penalty factor $\mu_t$, the updated policy in EPO is guided towards the feasible region and eventually converges to the optimal policy under certain conditions. Below, we present the theoretical guarantees.

\begin{lemma}
    \label{lemma1}  
    Suppose $\pi_t$ is the global maximum policy of the penalty function $P(\pi, \mu_t)$ with factor $\mu_t$, $\bar{\pi}$ is the optimal solution of the constrained problem (\ref{eq6}), and $F_R(\pi)$ is the objective function, then the inequality holds
    \begin{equation}
    \label{eq13}
        F_R(\bar{\pi}) \le P(\pi_t, \mu_t) \le F_R(\pi_t)
    \end{equation}
    
    Decreasing the penalty factor $\mu_{t+1} < \mu_t$, we get
    \begin{equation}
    \label{eq14}
    \begin{aligned}
        P(\pi_{t+1}, \mu_{t+1}) &\le P(\pi_t, \mu_t) \\
        F_R(\pi_{t+1}) &\le F_R(\pi_t) \\
    \end{aligned}
    \end{equation}
    
    The constraint $F_C(\pi_t)$ is monotonically non-increasing with the decreasing $\mu_t$ and same $\alpha$ if constraint is violated
    \begin{equation}
    \label{eq:cost}
        F_C(\pi_{t+1}) \le F_C(\pi_t)
    \end{equation}
\end{lemma}
\emph{Proof.} See the supplemental material.

Lemma \ref{lemma1} demonstrates that when the policy $\pi_t$ violates constraints, the maximum value of the penalty function $P(\pi_t,\mu_t)$ is greater than or equal to the optimal value $F_R(\bar{\pi})$ of problem (\ref{eq6}). 
By decreasing the penalty factor from $\mu_t$ to $\mu_{t+1}$, both $P(\pi_{t+1}, \mu_{t+1})$ and $F_R(\pi_{t+1})$ approach the solution $F_R(\bar{\pi})$, and the policy approaches the feasible region with a monotonically non-increasing constraint.
Furthermore, we establish the conditions ensuring that the updated policy in the penalty function satisfies the constraints.

\begin{theorem}
    \label{theorem1}
    Define a decreasing sequence of penalty factors $\{\mu_t\}$, suppose that the objective function $F_R(\pi)$ has an upper bound and $\pi_t$ is the global maximum policy of penalty function $P(\pi, \mu_t)$ with factor $\mu_t$. When $\mu_t \rightarrow 0$, the corresponding limit $\pi^*$ of the sequence $\{\pi_t\}$ is a solution of the constrained problem (\ref{eq6}). 
\end{theorem}

\begin{corollary}
    \label{corollary1}
    Consider the weight factor $\alpha=1$, $\pi_t$ is the global maximum policy of penalty function $P(\pi, \mu_t)$ with factor $\mu_t$, $\lambda^*$ is the Lagrange multiplier corresponding to the optimum $\bar{\pi}$ of the primal problem. When $\mu_t < \frac{1}{\parallel\lambda^*\parallel_\infty}$, $\pi_t$ is the solution of the constrained problem (\ref{eq6}).
\end{corollary}

\emph{Proof.} See the supplemental material.

According to Theorem \ref{theorem1}, we need to solve the sequential subproblems with decreasing $\mu_t$ to obtain the optimal solution of the problem (\ref{eq6}).
As $\mu_t$ decreases, policy $\pi_t$ converges to the optimal policy.
Corollary \ref{corollary1} guides us to raise $\alpha$ to 1 within the near-boundary region, guaranteeing the policy returns to the feasible region with a finite penalty factor.

\begin{algorithm}[h]
\caption{EPO: Exterior Penalty Policy Optimization} \label{alg1}
\hspace*{0.02in} {\bf Input:} 
Initialize policy network $\pi_\theta$, value networks $V_R^\omega$, $V_N^\upsilon$ and $V_F^\psi$. Set the hyperparameter for PPO clip rate, $\mu$, $\alpha$ for penalty function and learning rate $\eta$.\\
\hspace*{0.02in} {\bf Output:}
The optimal policy parameter $\theta$.

\begin{algorithmic}[1]
\label{algo1}
\FOR {epoch $k=0,1,2,...$}
\STATE Sample $N$ trajectories $\tau_{1}, ..., \tau_{N}$ under the current policy $\pi_{\theta_{k}}$. 
\STATE Process the trajectories to $C$-returns, and calculate advantage functions with $V_N^\upsilon$ and $V_F^\psi$.
\FOR{$K$ iterations}
\STATE Update value networks $V_R^\omega$, $V_N^\upsilon$ and $V_F^\psi$.
\STATE Update policy network $\pi_\theta$ with first-order optimizer $\theta_{k+1} = \theta_{k} +  \eta  \bigtriangledown_{\theta}  \left( L_R^{CLIP}(\theta)-\Psi(\mu)\Phi(log(1+e^{L_C^{CLIP}(\theta)})) \right)$.
\IF{ $\frac{1}{N}\sum_{j=1}^N D_{KL}(\pi_\theta||\pi_{\theta_k})[s_j]>\delta$}
\STATE Break.
\ENDIF
\ENDFOR
\STATE Decrease $\mu_{t+1} \in (0, \mu_t)$ and adjust $\alpha \in [0, 1]$ based on $F_C(\pi)$.
\ENDFOR
\RETURN policy parameters $\theta=\theta_{k+1}$.
\end{algorithmic}
\end{algorithm}

\subsection{Surrogate Penalty Function within Trust Region and Theoretical Bounds}

The complex dependency of state visitation distribution $d^\pi(s)$ on the unknown policy $\pi$ makes (\ref{eq9}) difficult to optimize directly. In this paper, we use samples generated by the current policy $\pi_k$ to locally approximate the original problem (\ref{eq6}) within the trust region:

\begin{equation}
\label{eq30}
\begin{aligned}
    \pi_{k+1}=& \arg\max \limits_{\pi \in {\Pi_\theta}} \mathbb{E}_{s\sim{d^{\pi_k}},a\sim\pi} [A_R^{\pi_k}(s,a)]  \\
    s.t. \quad & J_C(\pi_k)+ \frac{1}{1-\gamma} \mathbb{E}_{s\sim{d^{\pi_k}},a\sim\pi} [A_C^{\pi_k}(s,a)] \leq d \\
    & D(\pi\|\pi_k) \leq \delta
\end{aligned}
\end{equation}
where $\Pi_\theta$ represents the policy set parameterized by parameter $\theta$, $D(\pi\|\pi_k)=\mathbb{E}_{s\sim{\pi_k}}[D_{KL(\pi\|\pi_k)}[s]]$, $D_{KL}$ is the KL divergence, and $\delta>0$ is the step size. The set $\{\pi \in \Pi_\theta:D(\pi\|\pi_k) \leq \delta\}$ is the trust region.
The surrogate objective and constraint are defined in a simplified formulation:

\begin{equation}
\label{eq31}
\begin{aligned}
    &L_R^{\pi_k}(\pi) := \mathbb{E}_{s\sim{d^{\pi_k}},a\sim\pi} [A_R^{\pi_k}(s,a)]  \\
    &L_C^{\pi_k}(\pi) := \frac{1}{1-\gamma} \mathbb{E}_{s\sim{d^{\pi_k}},a\sim\pi} [A_C^{\pi_k}(s,a)] + J_C(\pi_k) - d \\
\end{aligned}
\end{equation}

Then the surrogate penalty objective function is:

\begin{equation}
\label{eq36}
\begin{aligned}
    \arg\max \limits_{\pi \in {\Pi_\theta}} & L_R^{\pi_k}(\pi)-\frac{1}{\mu}\left(\alpha L_C^{\pi_k+}(\pi) + (1-\alpha)(L_C^{\pi_k+}(\pi))^2 \right) \\
    s.t. \quad & D(\pi\|\pi_k) \leq \delta
\end{aligned}
\end{equation}

EPO is a trust region method, so it satisfies the worst-case bound for the updated policy in CPO.

\begin{proposition}[\cite{achiam2017constrained}]
    \label{proposition1}
    Suppose surrogate constraint $L_C^{\pi_k}(\pi_{k+1})$ is satisfied, the upper bound on constraint violation for each update is

    \begin{equation}
    \label{eq32}
    \begin{aligned}
        J_C(\pi_{k+1}) \le d + \frac{\sqrt{2\delta}\gamma\epsilon_C^{\pi_{k+1}}}{(1-\gamma)^2}
    \end{aligned}
    \end{equation}
    where $\epsilon_C^{\pi_{k+1}}:=\max_s|\mathbb{E}_{a\sim{\pi_{k+1}}}[A_C^{\pi_k}(s,a)]|$.
\end{proposition}

The local approximation approach introduces bias. We further give the worst-case error bound for the surrogate penalty function compared to the original penalty function.
We denote the surrogate penalty function for problem (\ref{eq30}) as $P^{\pi_k}(\pi, \mu_t)=L_R^{\pi_k}(\pi)-\frac{1}{\mu_t}\Phi(L_C^{\pi_k+}(\pi))$.

\begin{theorem}
    \label{theorem2}
    Suppose surrogate constraint $L_C^{\pi_k}(\pi)$ is satisfied, the upper bound on the approximation error between the surrogate and the original penalty functions is
    
    \begin{equation}
    \label{eq33}
    \begin{aligned}
        |P(\pi, &\mu_t)-P^{\pi_k}(\pi,\mu_t)| \le \frac{\sqrt{2\delta}\gamma\epsilon_R^\pi}{1-\gamma} + \\
        &\frac{1}{\mu_t}\left( \frac{\sqrt{2\delta}\alpha\gamma\epsilon_C^\pi}{(1-\gamma)^2} +
        \left(\frac{\sqrt{2\delta(1-\alpha)}\gamma\epsilon_C^\pi}{(1-\gamma)^2}\right)^2  \right)
    \end{aligned}
    \end{equation}
    where $\epsilon_R^\pi:=\max_s|\mathbb{E}_{a\sim\pi}[A_R^{\pi_k}(s,a)]|$, and $\epsilon_C^\pi:=\max_s|\mathbb{E}_{a\sim\pi}[A_C^{\pi_k}(s,a)]|$.
\end{theorem}

\emph{Proof.} See the supplemental material.

\subsection{Smooth Penalty Function in Practical Implementation}

The penalty function in (\ref{eq36}) is mostly smooth except at the constraint boundary. However, the constraint boundary is crucial in CRL, particularly when the optimal policy lies close to it. 
The first-order optimizer needs precise gradient information during policy updates.
To address these issues, we propose a novel smooth penalty function containing smoothing and factor terms in the implementation:

\begin{equation}
\label{eq25}
\begin{aligned}
    &P_S(\pi, \mu) := F_R(\pi) - \Psi(\mu)\Phi(log(1+e^{F_C(\pi)})) \\
    &where \quad \Psi(\mu) = \frac{1}{\mu}e^{F_C(\pi)}
\end{aligned}
\end{equation}
where $log$ denotes the natural logarithm, $log(1+e^{F_C(\pi)})$ is the smoothing term. $\Phi(x) = \alpha x + (1-\alpha) x^2$ represents the penalty term, $\Psi(\mu)$ is the penalty factor term.

The smooth term ensures that the penalty function remains differentiable throughout the parameter space.
The adaptive factor function $\Psi(\mu)$ serves as a strategy for determining the penalty factor based on the empirical value of $F_C$, and it remains detached from the gradient calculation. This effectively reduces the frequency of factor adjustments, thus decreasing the number of sub-problems.
For this implementation version in (\ref{eq25}), Lemma \ref{lemma1} still guarantees monotonically non-increasing constraints, and Theorem \ref{theorem1} ensures that the practical policy converges to the feasible region. The proof is shown in the supplemental material.

To address the divergence constraint in the surrogate objective function (\ref{eq30}), we employ the clipped surrogate function of the PPO \cite{schulman2017proximal} for approximation. Then, we get the final penalty objective in practice:

\begin{equation}
\label{eq168}
\begin{aligned}
    \arg\max \limits_{\pi \in {\Pi_\theta}}  &L_R^{CLIP}(\pi)-\Psi(\mu)\Phi(log(1+e^{L_C^{CLIP}(\pi)}))
\end{aligned}
\end{equation}

\begin{equation}
\label{eq166}
\begin{aligned}
    &L_R^{CLIP}(\pi)=\mathbb{E}_{s\sim{d^{\pi_k}},a\sim{\pi_k}}[min\{ r(\pi)A_R^{\pi_k}(s,a), 
    \\ &\quad clip(r(\pi), 1-\epsilon,1+\epsilon)A_R^{\pi_k}(s,a) \}] \\
    &L_C^{CLIP}(\pi)=\frac{1}{1-\gamma}\mathbb{E}_{s\sim{d^{\pi_k}},a\sim{\pi_k}}[min\{r(\pi)A_C^{\pi_k}(s,a), \\
    &\quad clip(r(\pi), 1-\epsilon,1+\epsilon)A_C^{\pi_k}(s,a)\}] +J_C(\pi_k)-d
\end{aligned}
\end{equation}
where $r(\pi)= \frac{\pi}{\pi_k}$ is the importance sampling ratio, $\epsilon$ is the clip ratio. Algorithm \ref{algo1} presents a summary of EPO. The implementation code of EPO is available at \url{https://github.com/Ontroad/EPOPMN}.

\section{Experiment}

\begin{figure*}[ht]
  \centering
  \begin{subfigure}{0.245\textwidth}
    \includegraphics[width=\linewidth]{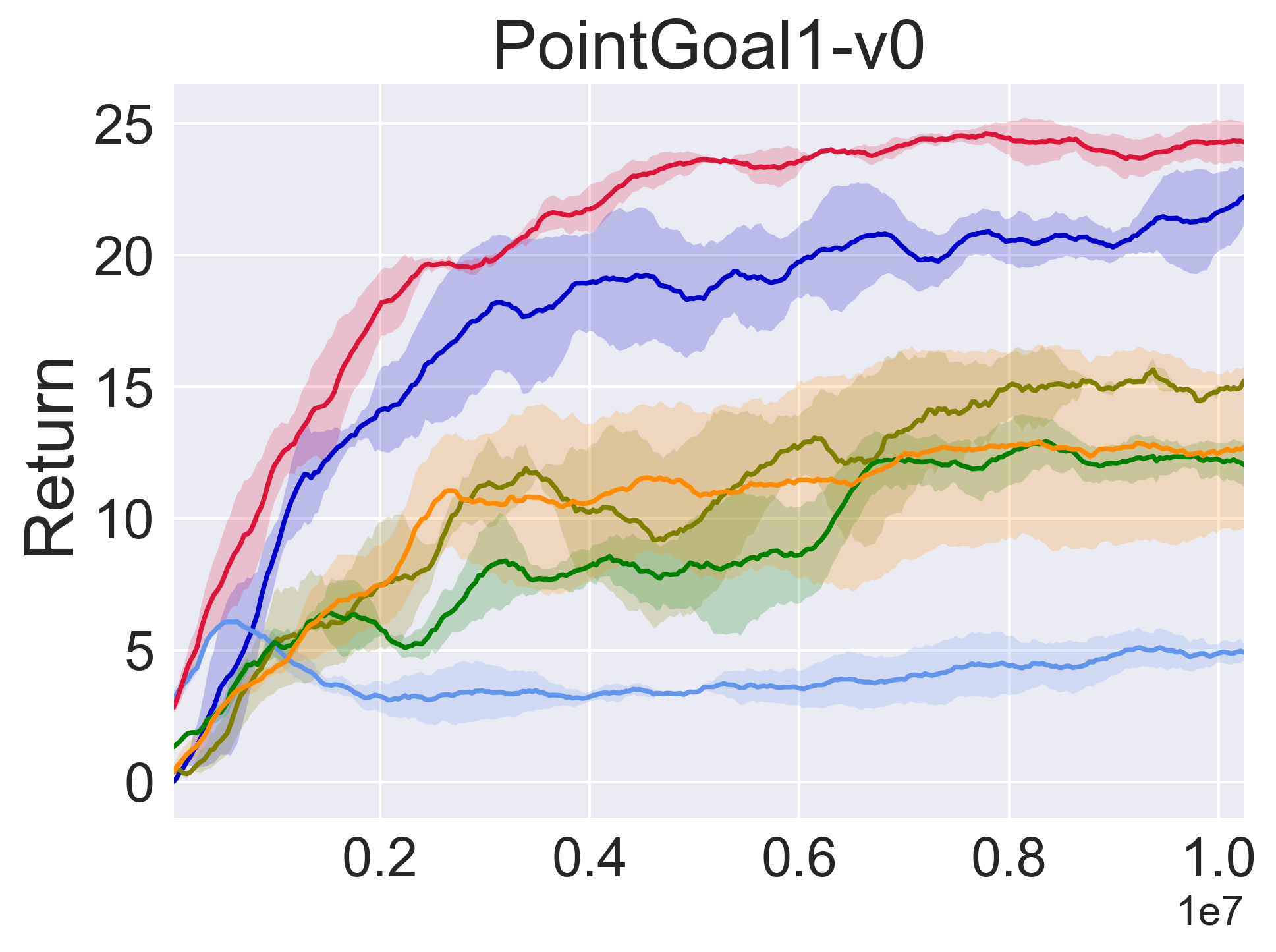}
    \label{fig:img10}
  \end{subfigure}
  \hfill
  \begin{subfigure}{0.245\textwidth}
    \includegraphics[width=\linewidth]{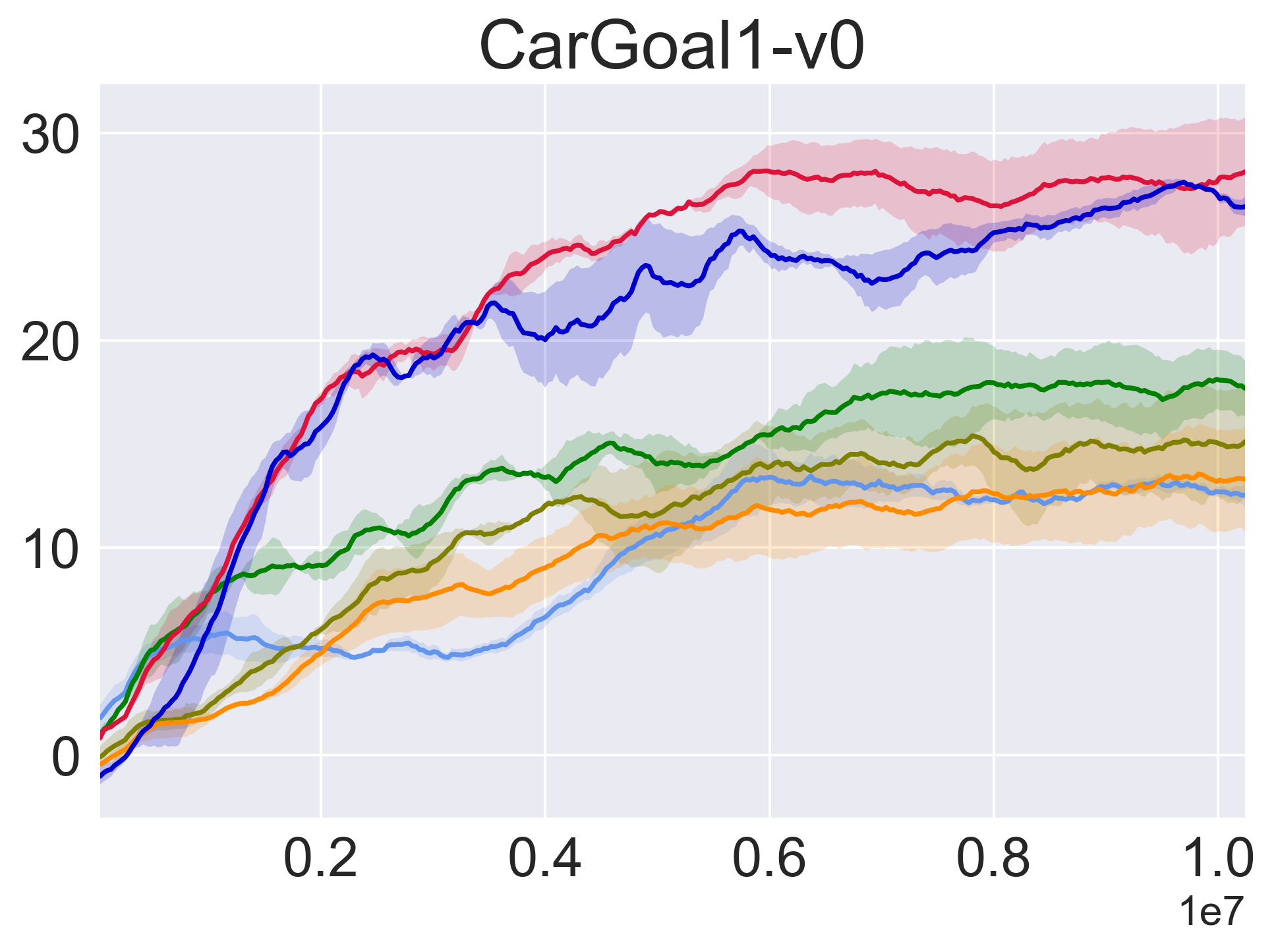}
    \label{fig:img11}
  \end{subfigure}
  \hfill
  \begin{subfigure}{0.245\textwidth}
    \includegraphics[width=\linewidth]{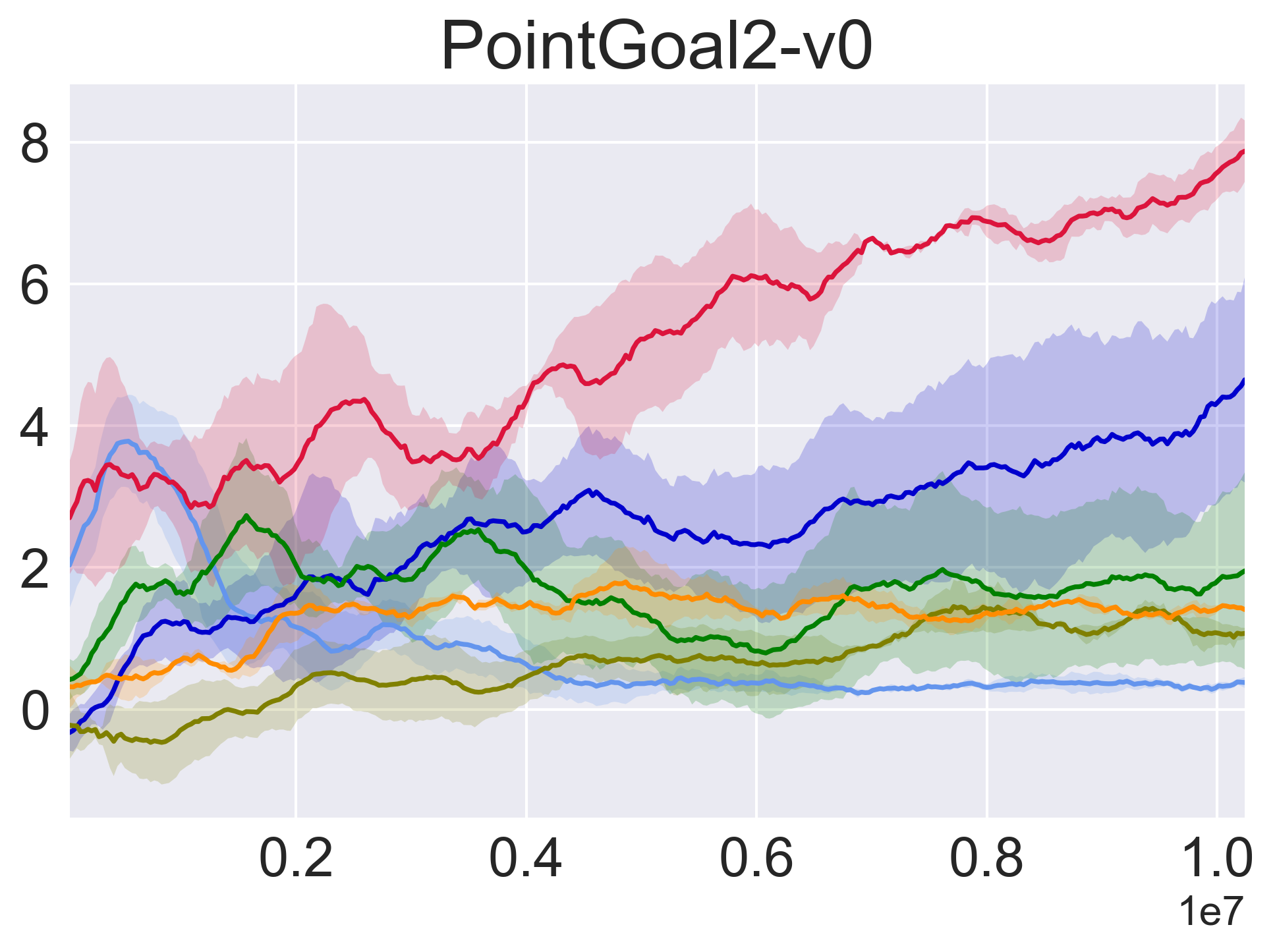}
    \label{fig:img12}
  \end{subfigure}
  \hfill
  \begin{subfigure}{0.245\textwidth}
    \includegraphics[width=\linewidth]{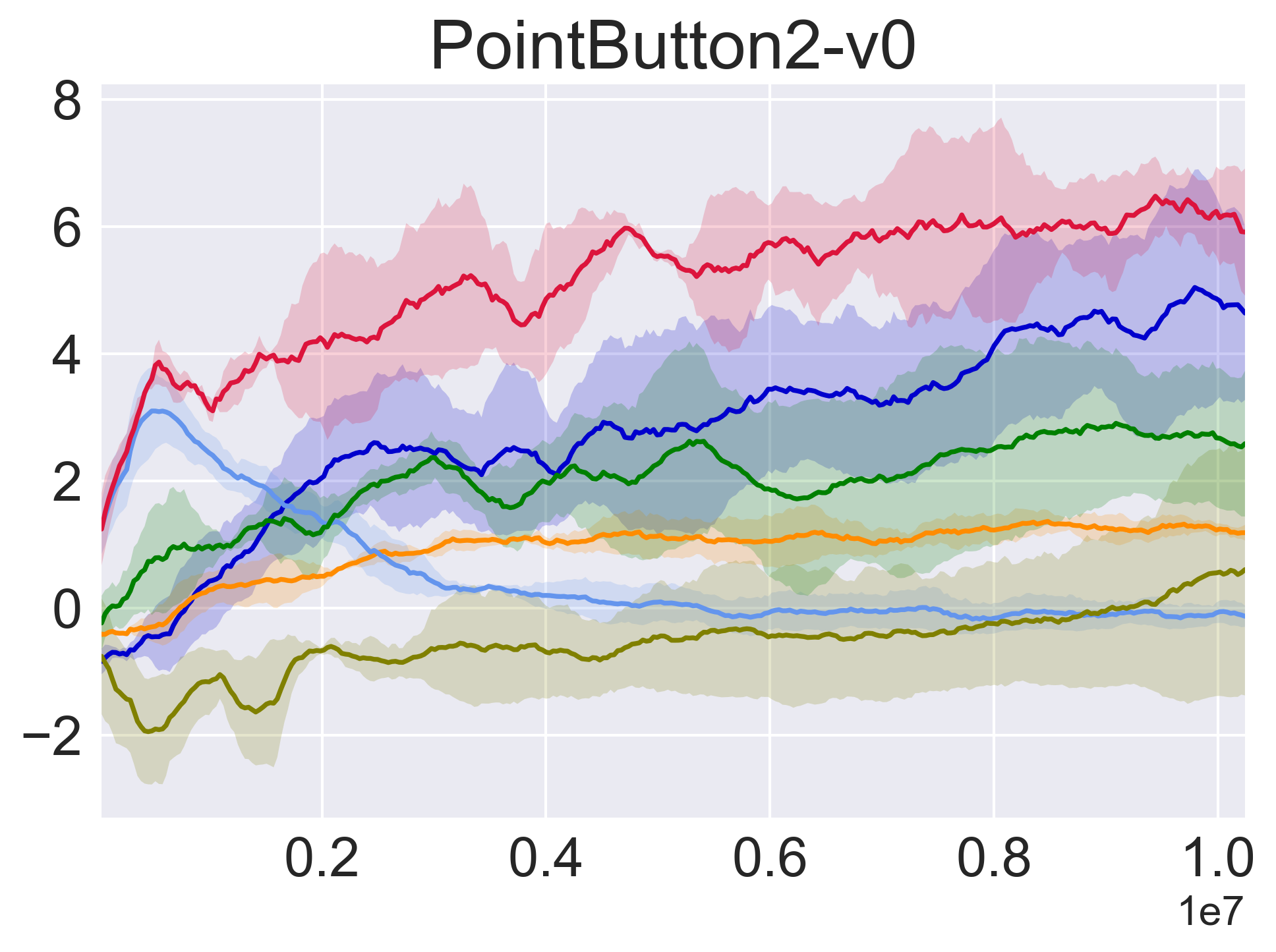}
    \label{fig:img13}
  \end{subfigure} 
  
  \begin{subfigure}{0.245\textwidth}
    \includegraphics[width=\linewidth]{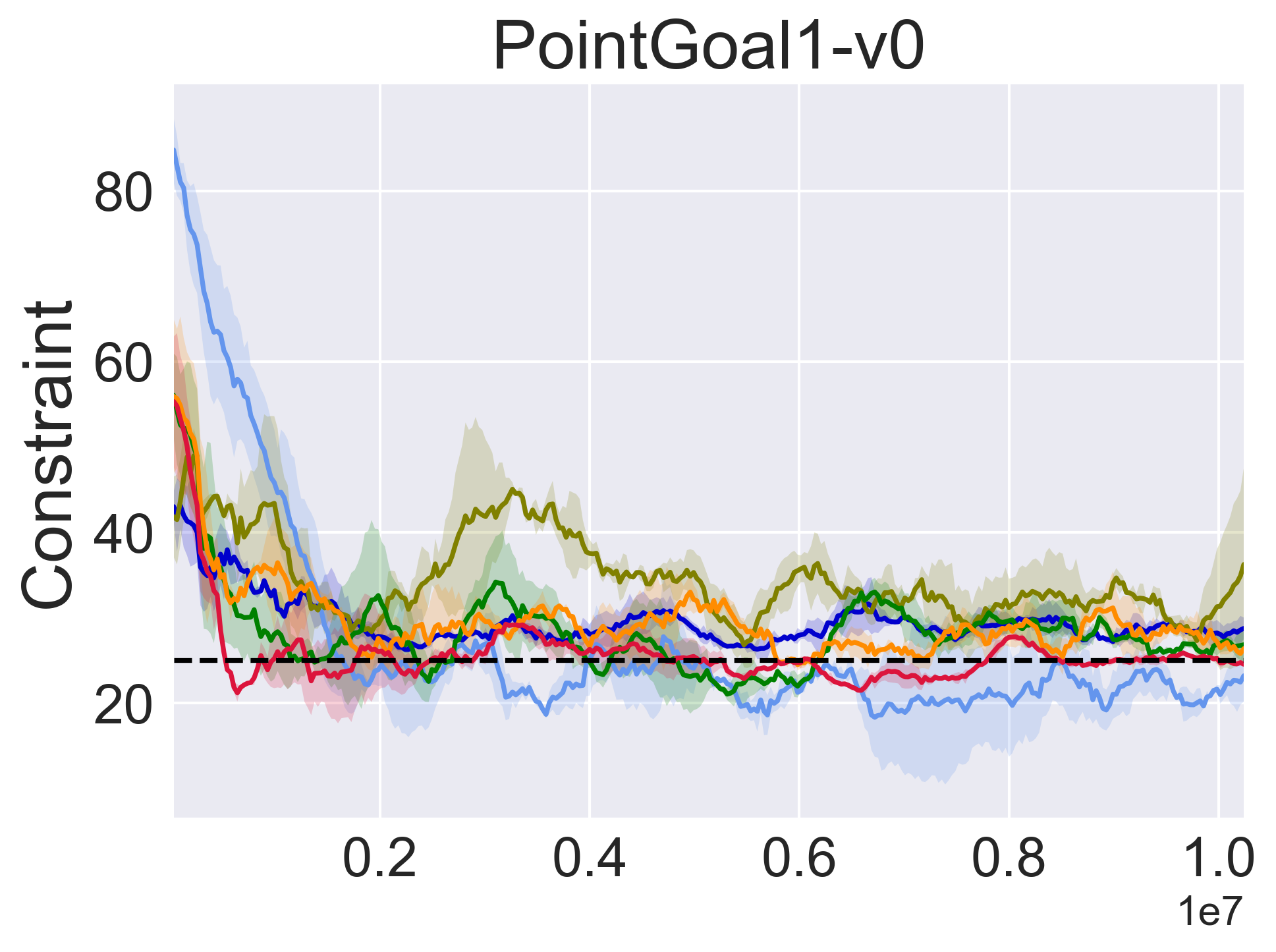}
    \label{fig:img14}
  \end{subfigure}
  \hfill
  \begin{subfigure}{0.245\textwidth}
    \includegraphics[width=\linewidth]{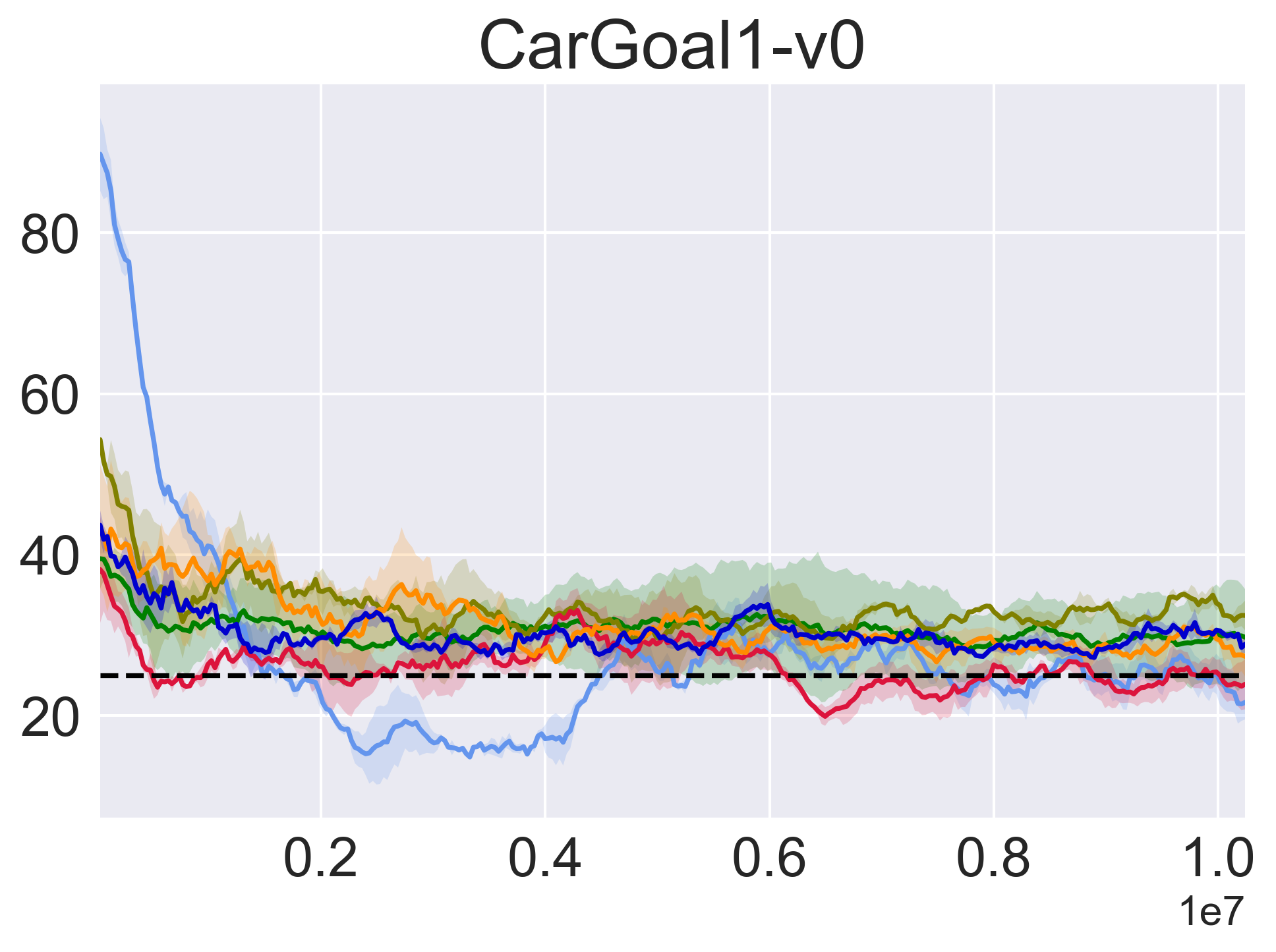}
    \label{fig:img15}
  \end{subfigure}
  \hfill
  \begin{subfigure}{0.245\textwidth}
    \includegraphics[width=\linewidth]{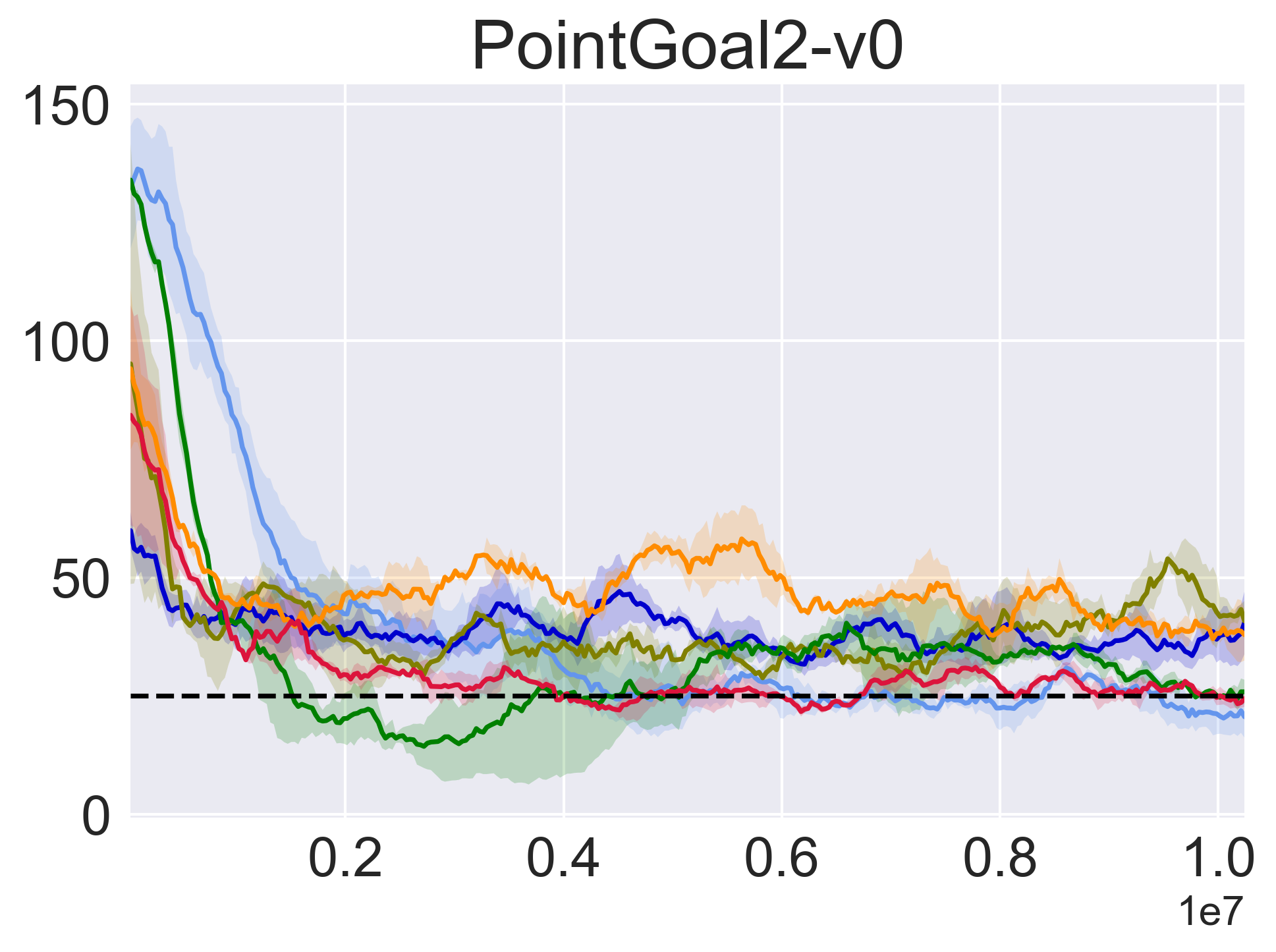}
    \label{fig:img16}
  \end{subfigure}
  \hfill
  \begin{subfigure}{0.245\textwidth}
    \includegraphics[width=\linewidth]{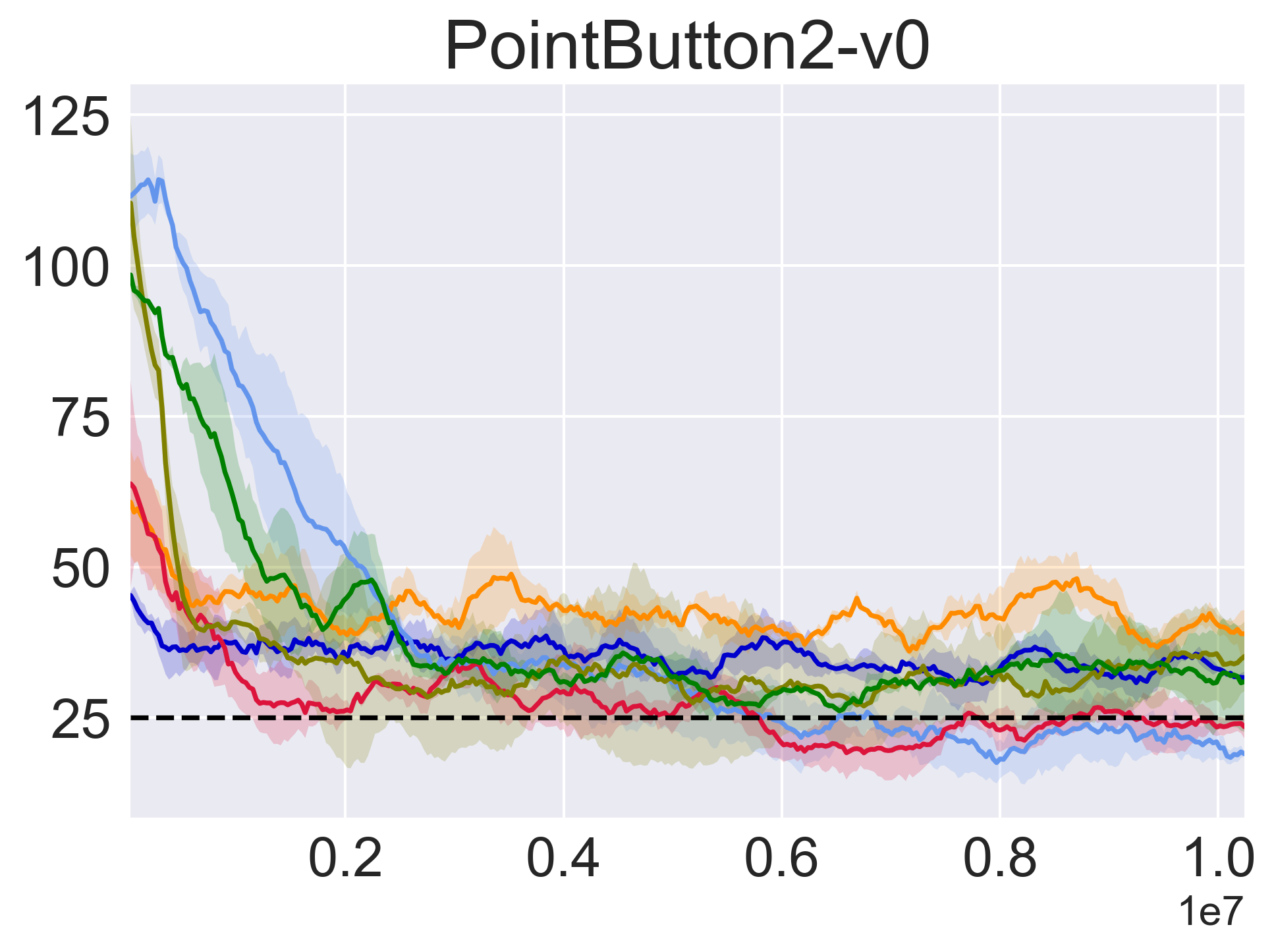}
    \label{fig:img17}
  \end{subfigure}

  \begin{subfigure}{0.5\textwidth}
    \includegraphics[width=\linewidth]{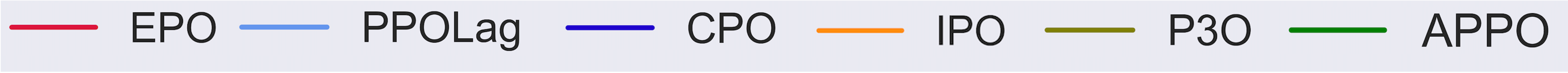}
    \label{fig:img18}
  \end{subfigure}
  \hfill
  \caption{Comparison of EPO to baselines over 3 seeds on Safety Gymnasium. The x-axis is the training steps, the y-axis is the average return or constraint. The solid line is the mean and the shaded area is the standard deviation. The dashed line marks the constraint limit of 25.}
  \label{fig:gym}
\end{figure*}

\begin{figure*}[ht]
  \centering
  \begin{subfigure}{0.245\textwidth}
    \includegraphics[width=\linewidth]{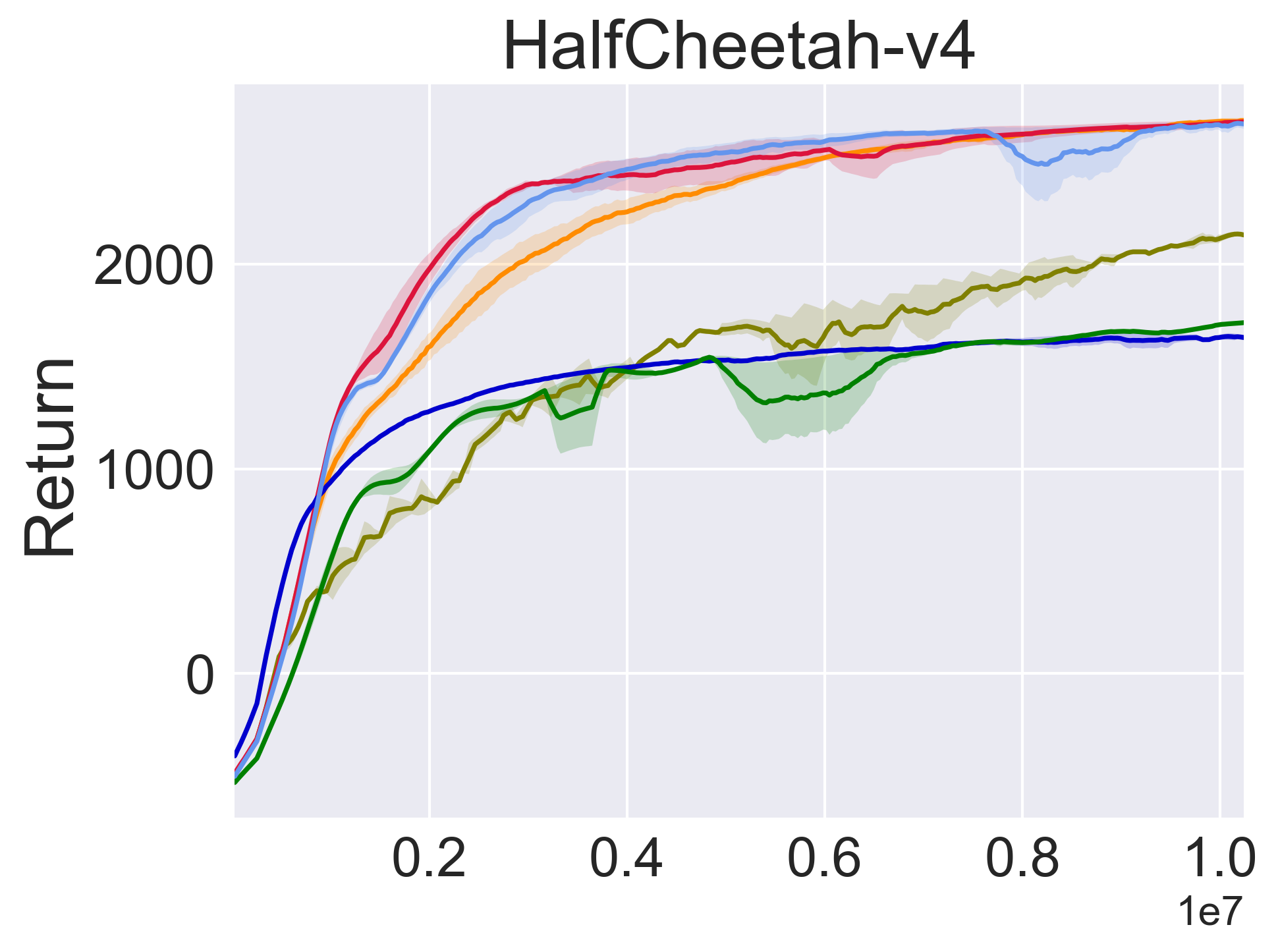}
    \label{fig:img1}
  \end{subfigure}
  \hfill
  \begin{subfigure}{0.245\textwidth}
    \includegraphics[width=\linewidth]{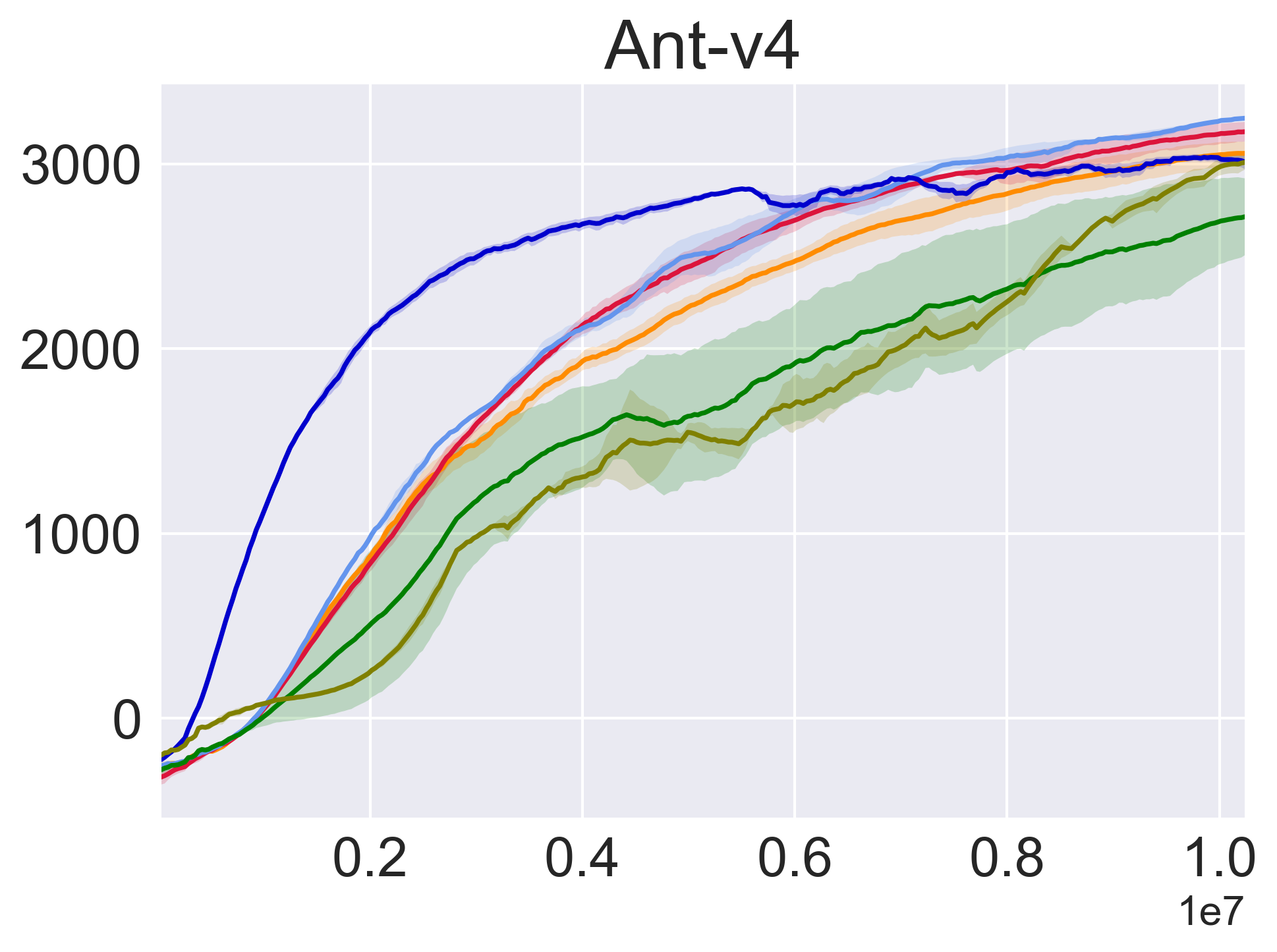}
    \label{fig:img2}
  \end{subfigure}
  \hfill
  \begin{subfigure}{0.245\textwidth}
    \includegraphics[width=\linewidth]{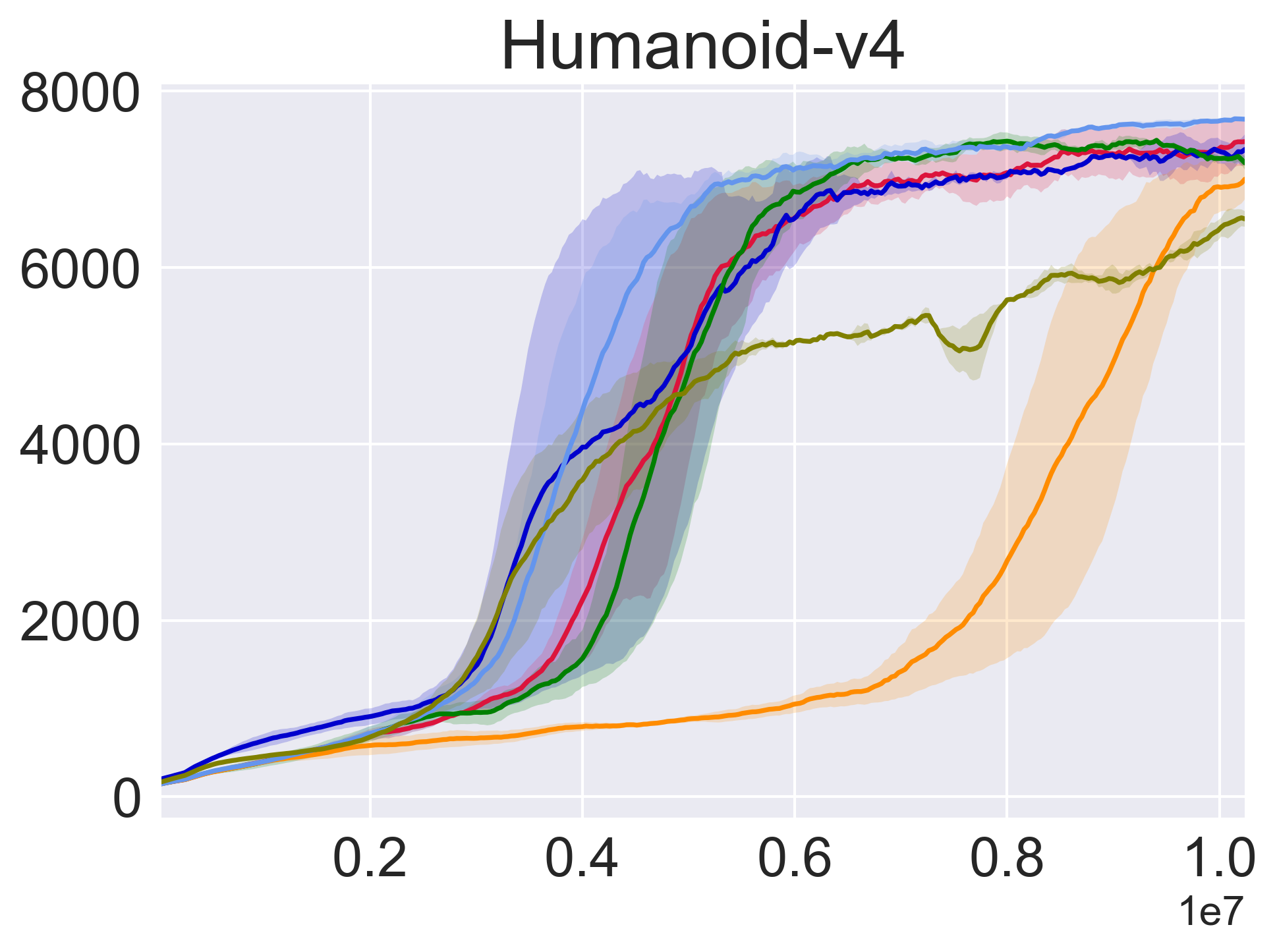}
    \label{fig:img3}
  \end{subfigure}
  \hfill
  \begin{subfigure}{0.245\textwidth}
    \includegraphics[width=\linewidth]{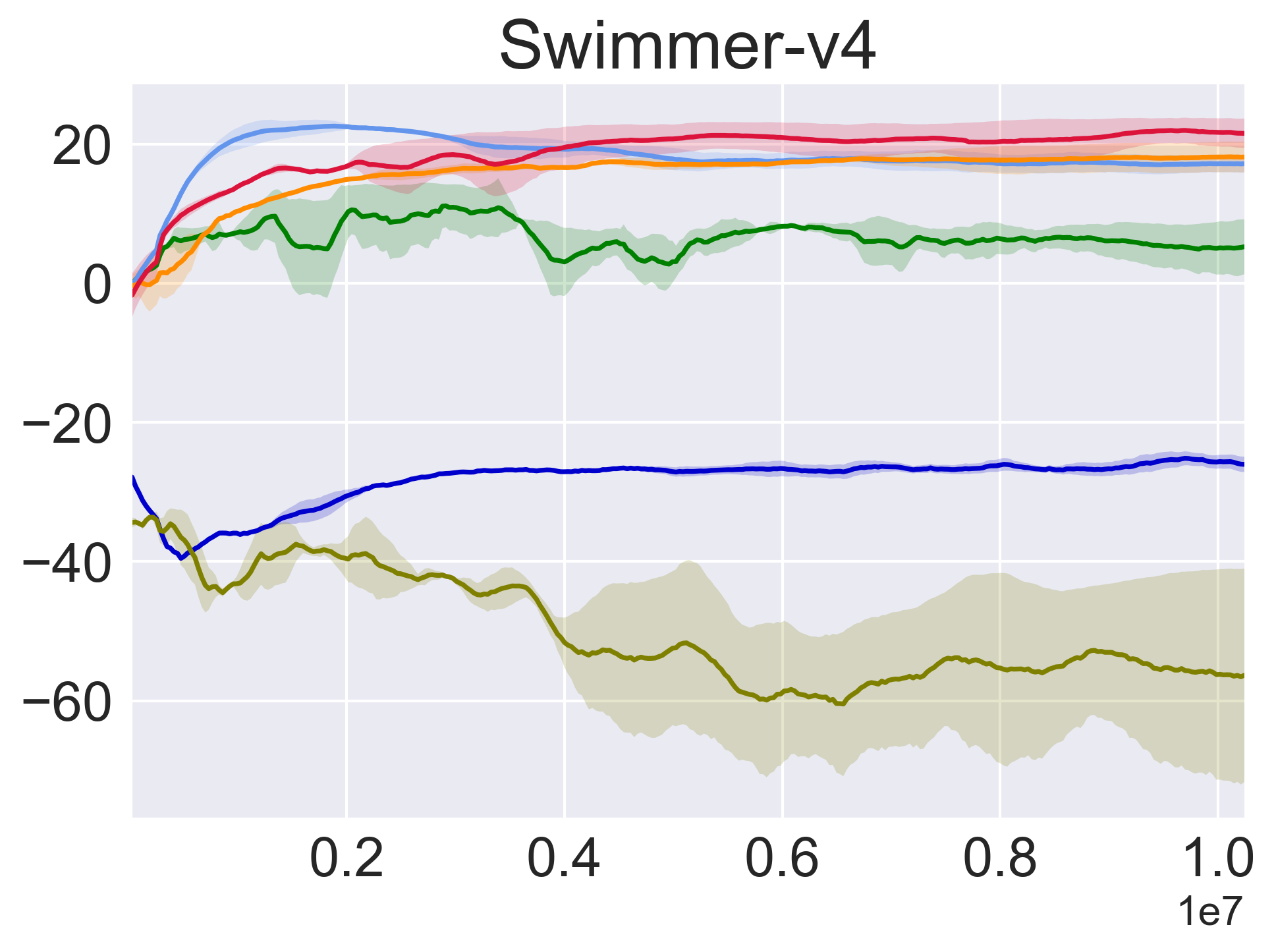}
    \label{fig:img4}
  \end{subfigure}
  
  \begin{subfigure}{0.245\textwidth}
    \includegraphics[width=\linewidth]{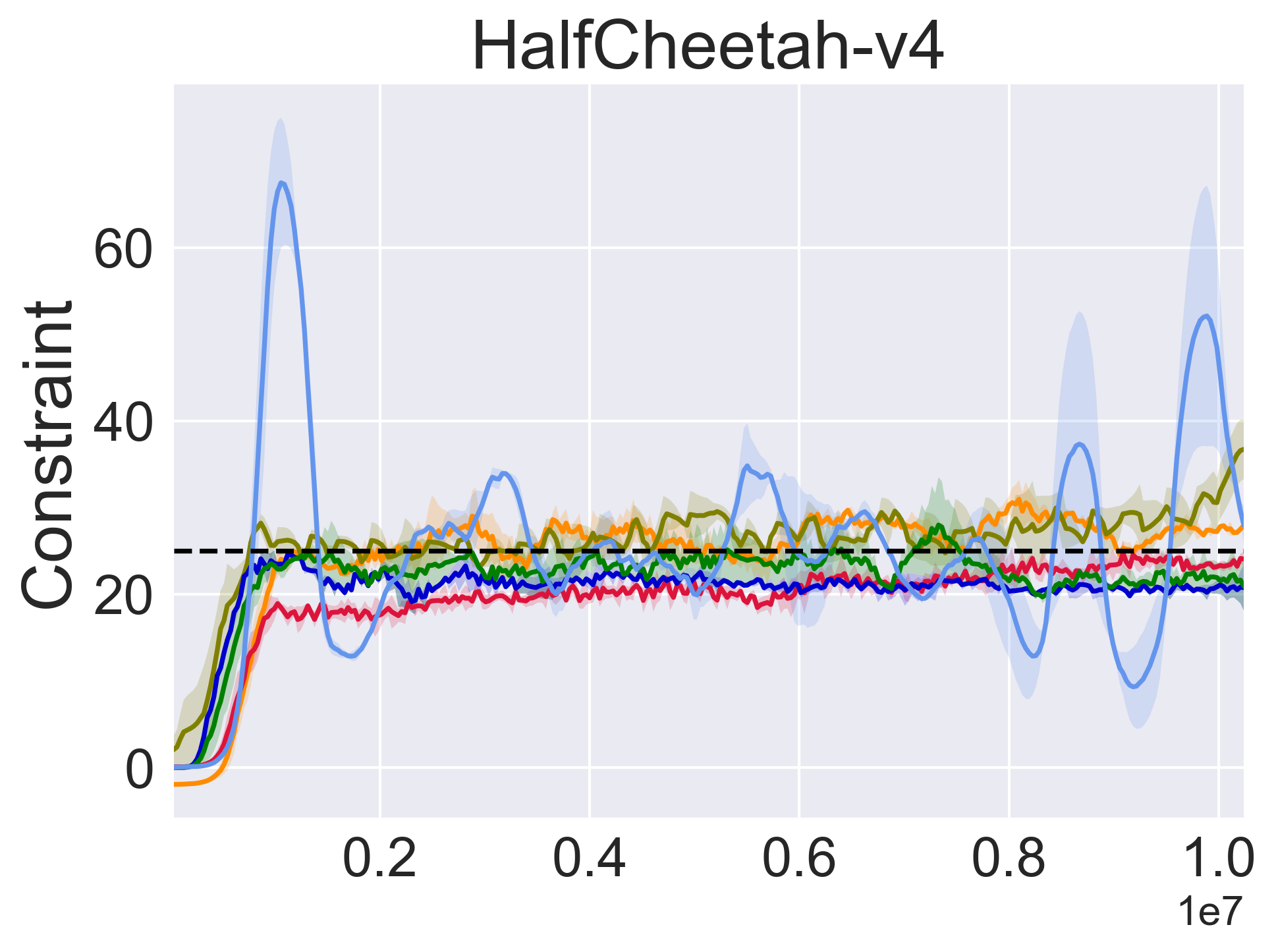}
    \label{fig:img5}
  \end{subfigure}
  \hfill
  \begin{subfigure}{0.245\textwidth}
    \includegraphics[width=\linewidth]{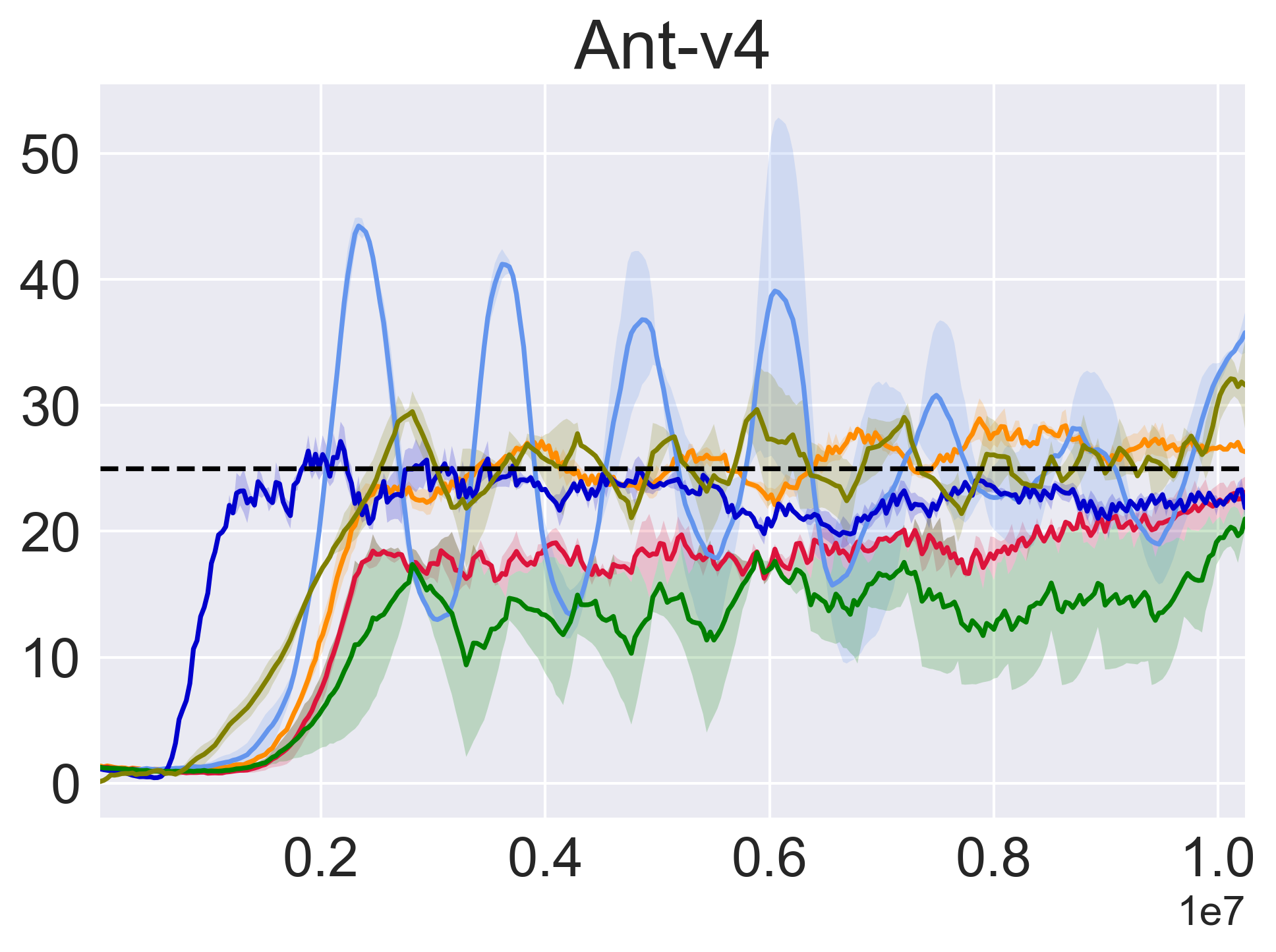}
    \label{fig:img6}
  \end{subfigure}
  \hfill
  \begin{subfigure}{0.245\textwidth}
    \includegraphics[width=\linewidth]{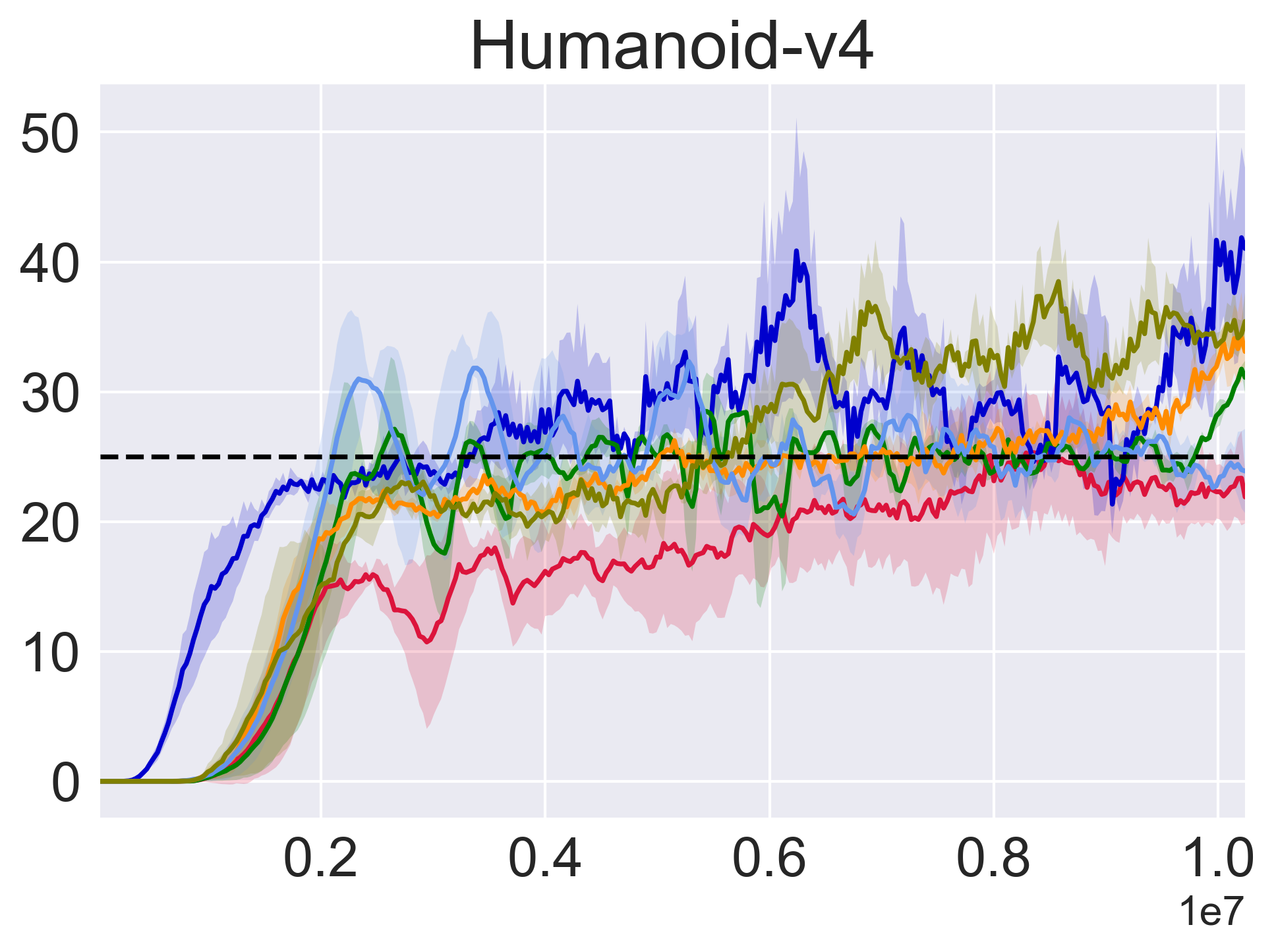}
    \label{fig:img7}
  \end{subfigure}
  \hfill
  \begin{subfigure}{0.245\textwidth}
    \includegraphics[width=\linewidth]{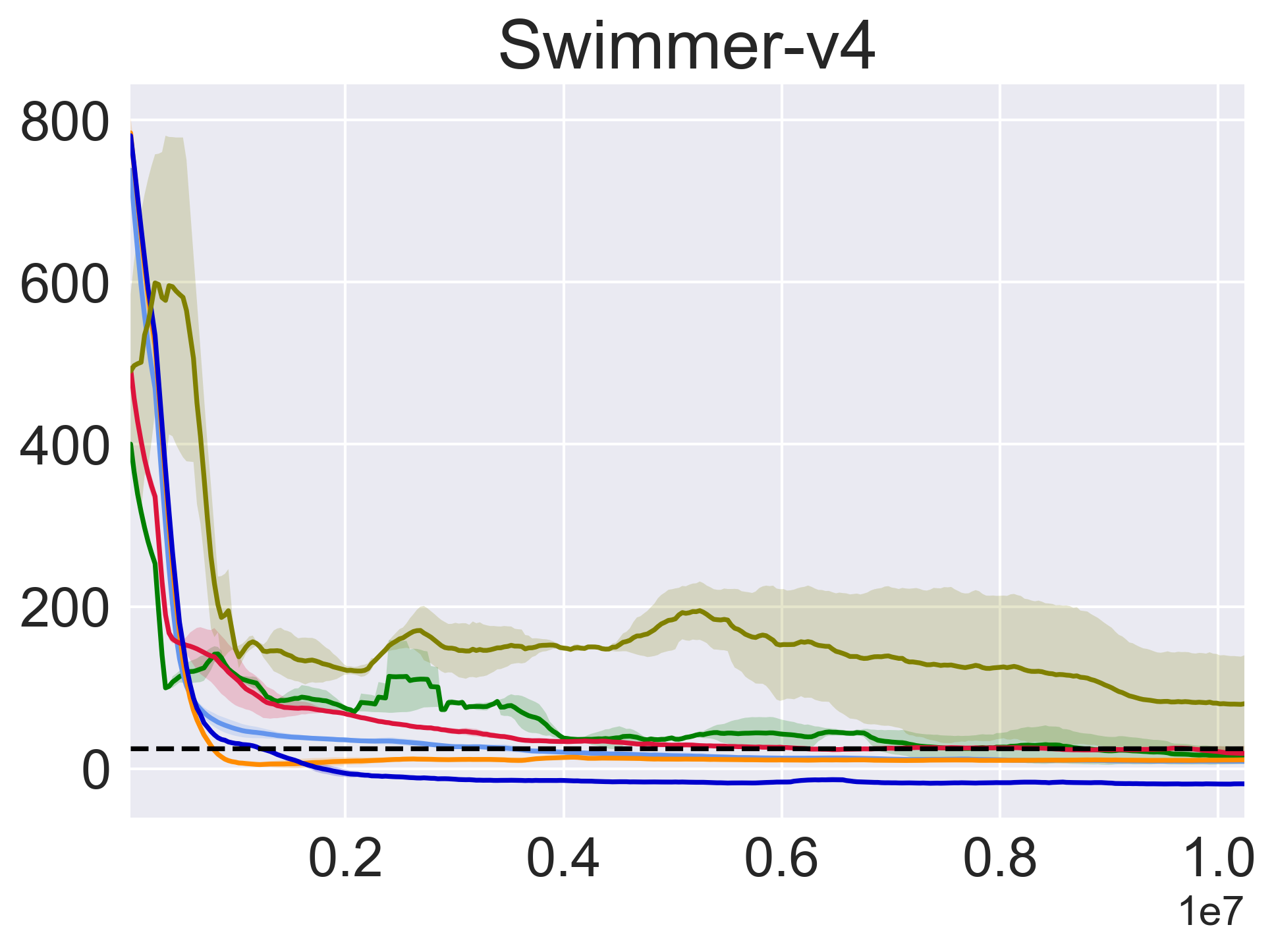}
    \label{fig:img8}
  \end{subfigure}
  
  \begin{subfigure}{0.5\textwidth}
    \includegraphics[width=\linewidth]{figures/legend.png}
    \label{fig:img9}
  \end{subfigure}
  \hfill
  \caption{Comparison of EPO to the baselines over 3 seeds on Safety MuJoCo.
  The dashed line marks the constraint limit of 25.}
  \label{fig:mujoco}
\end{figure*}

We train different agents and design comparison experiments in four navigation tasks based on Safety Gymnasium \cite{brockman2016openai} and four MuJoCo physical simulator tasks \cite{todorov2012mujoco}. Baselines include the primal-dual method PPO Lagrangian \cite{chow2017risk,schulman2017proximal}, the second-order optimization method CPO \cite{achiam2017constrained}, the augmented Lagrangian method APPO \cite{dai2023augmented}, and the penalty function method IPO \cite{liu2020ipo} and P3O \cite{zhang2022penalized}. 
For fairness of the results, we conduct multiple comparison experiments in the same environment and with the same settings.
Extensive experiments demonstrate that (1) EPO outperforms current state-of-the-art algorithms on policy performance and constraint satisfaction, and (2) provides stable convergence and fewer constraint violations during training.

\paragraph{Scenario Description.} All tasks aim to maximize the expected reward (the higher, the better) while satisfying the constraint (the lower, the better).
In Safety Gymnasium, we train different agents, including Point and Car, for various navigation tasks: (1) the Goal task, where agents navigate to the goal location while circumventing hazards, and (2) the Button task, where agents navigate to the goal button and touch the correct goal button while avoiding gremlins and hazards. Additionally, we test the more difficult Goal2 and Button2 tasks, requiring avoiding more hazards and vases, which have not been proven to be effectively solved by baselines.
In Safety MuJoCo, different agents are rewarded for running along a straight line with a velocity limit for safety and stability.

\begin{figure}[h]
    \centering
    \begin{subfigure}[t]{0.47\columnwidth}
         \centering
         \includegraphics[width=\textwidth]{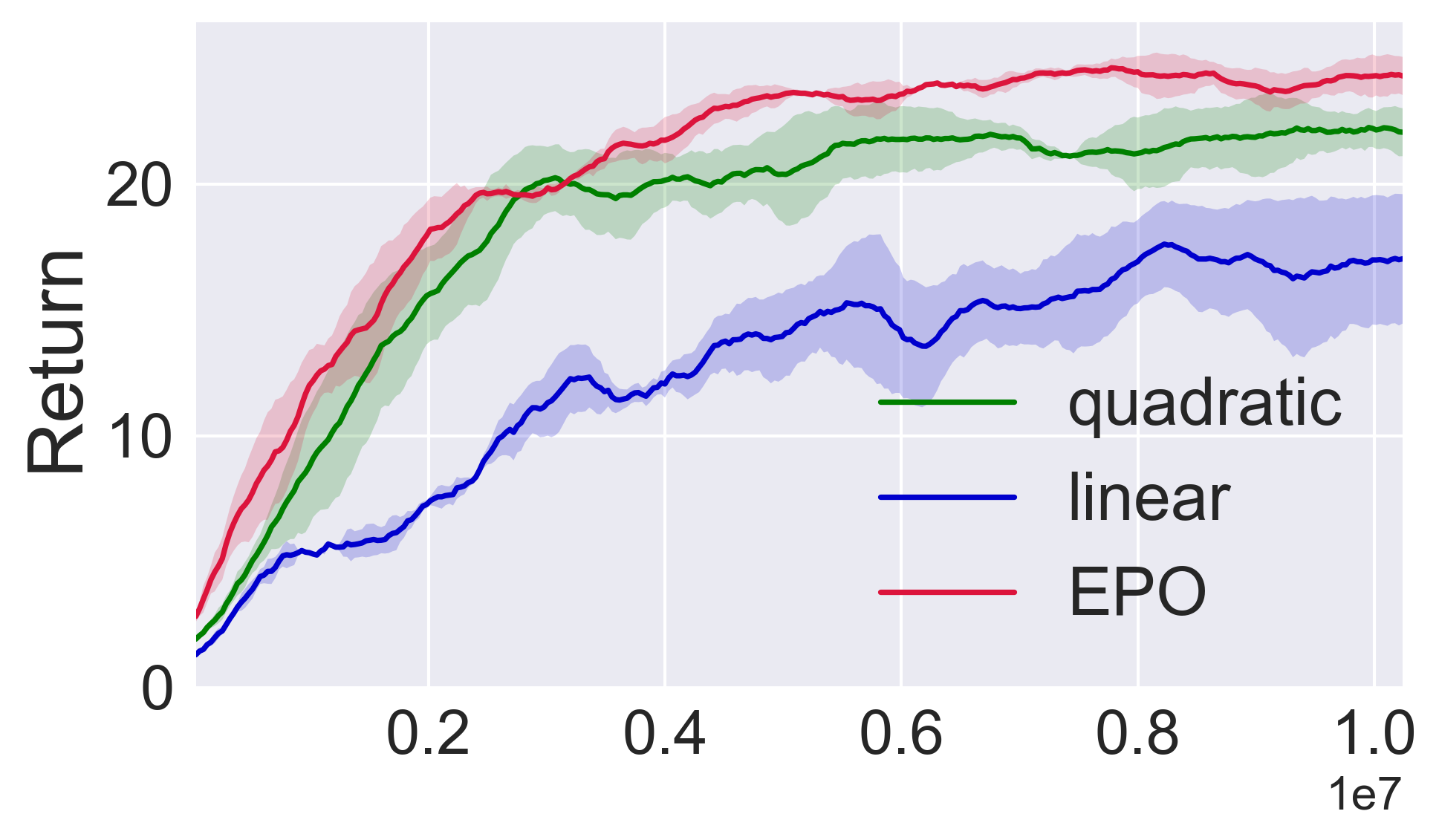}
        \caption{Return}
        \label{figa}
     \end{subfigure}
     \hfill
     \begin{subfigure}[t]{0.47\columnwidth}
         \centering
         \includegraphics[width=\textwidth]{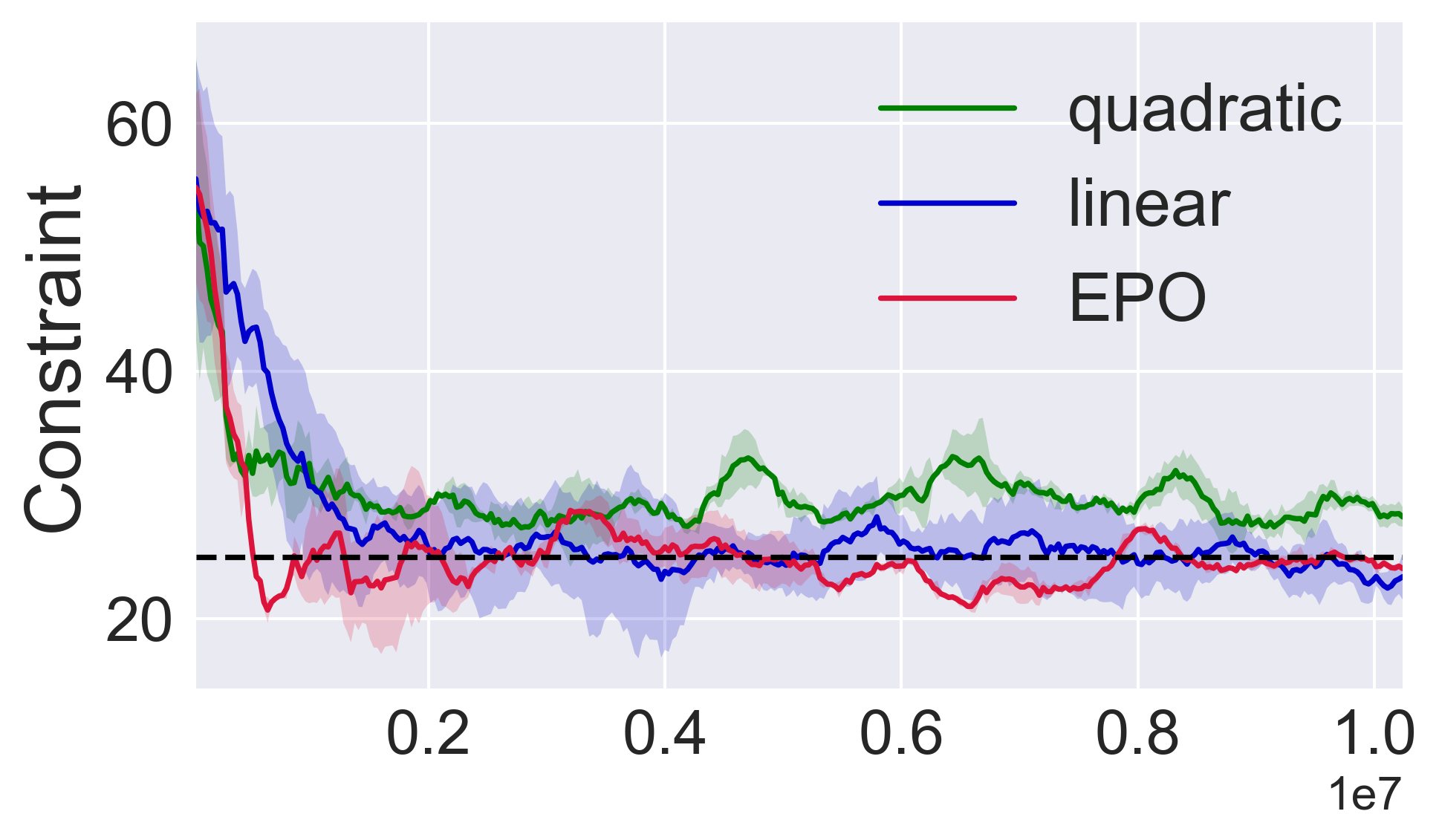}
        \caption{Constraint}
        \label{figb}
     \end{subfigure}
     \hfill
     
     \begin{subfigure}[t]{0.47\columnwidth}
         \centering
         \includegraphics[width=\textwidth]{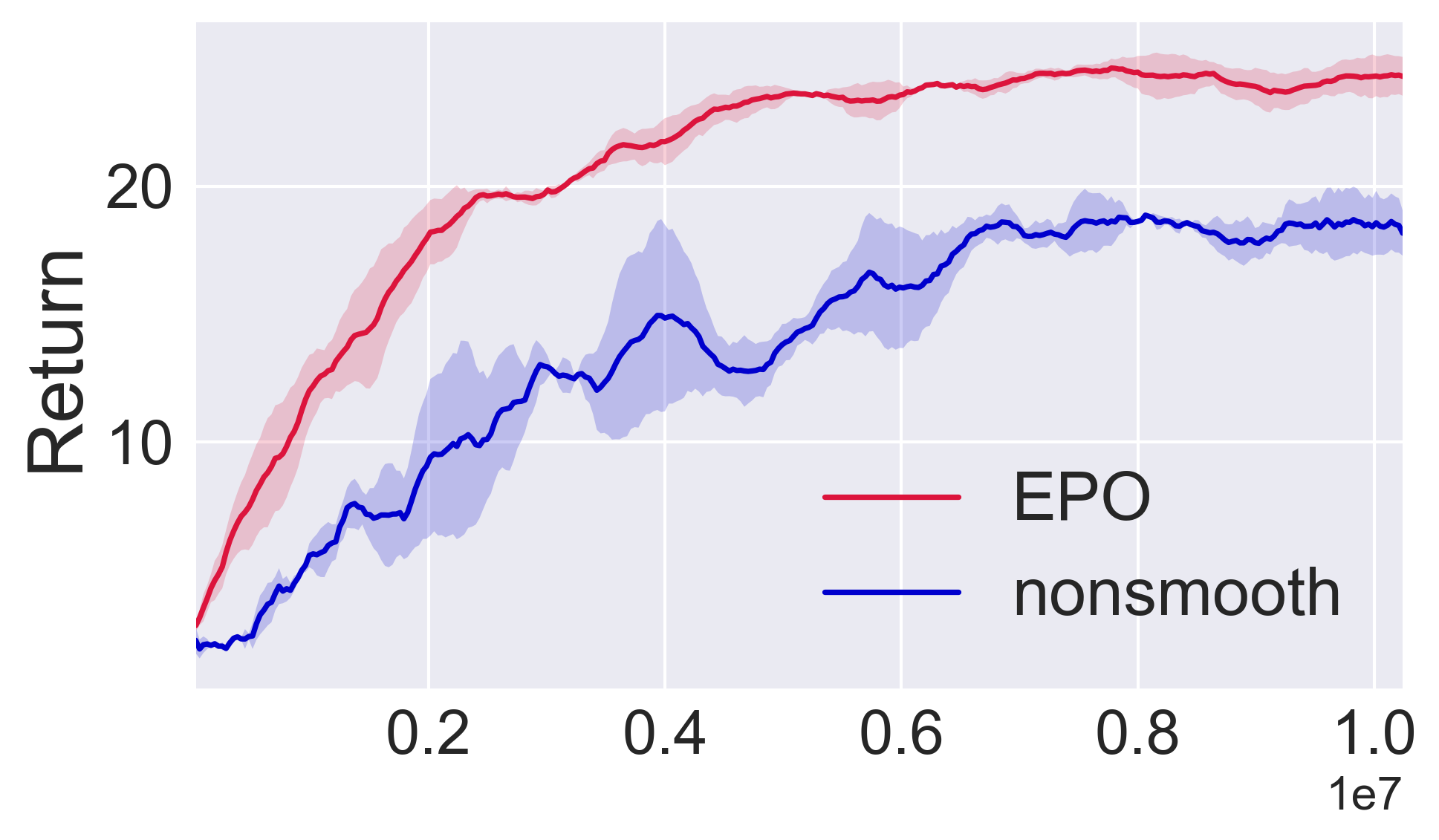}
         \caption{Return}
         \label{figc}
     \end{subfigure}
     \hfill
     \begin{subfigure}[t]{0.47\columnwidth}
         \centering
         \includegraphics[width=\textwidth]{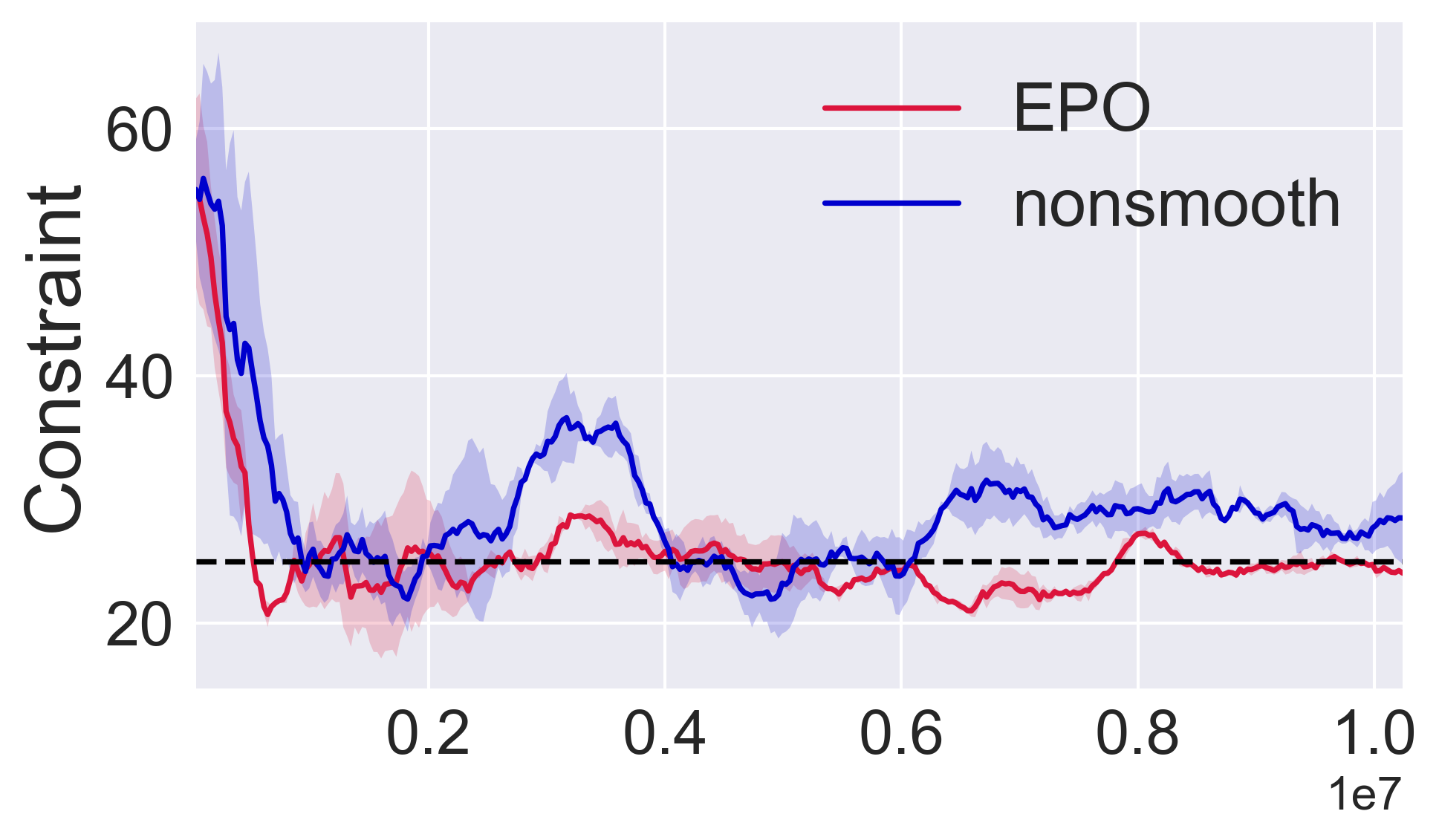}
         \caption{Constraint}
         \label{figd}
     \end{subfigure}
     \hfill
    \caption{Ablation experiments in PointGoal1-v0.}
    \label{fig:ablation}
\end{figure}

\begin{figure}[ht]
    \centering
    \begin{subfigure}[t]{0.47\columnwidth}
         \centering
         \includegraphics[width=\textwidth]{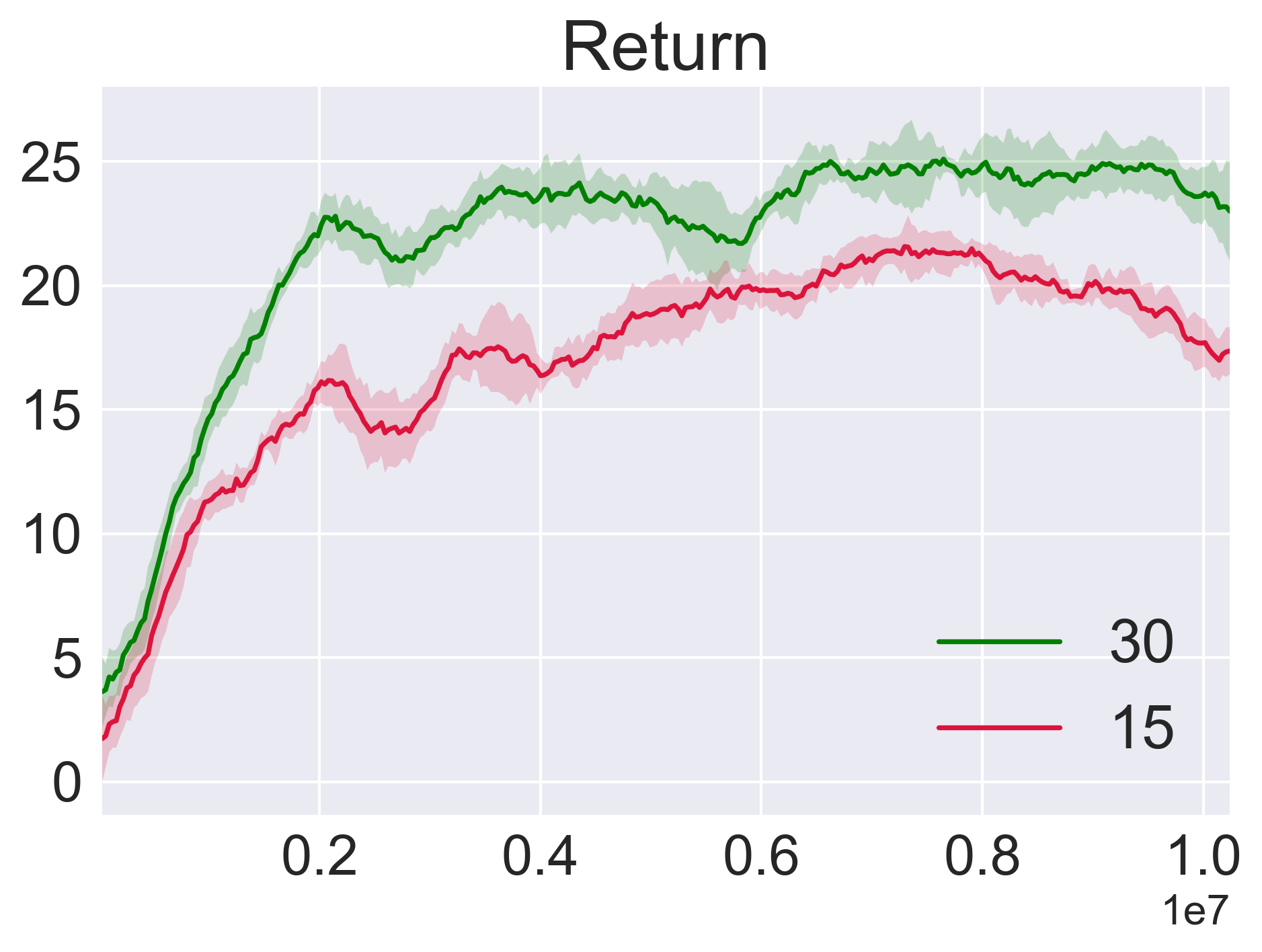}
     \end{subfigure}
     \hfill
     \begin{subfigure}[t]{0.47\columnwidth}
         \centering
         \includegraphics[width=\textwidth]{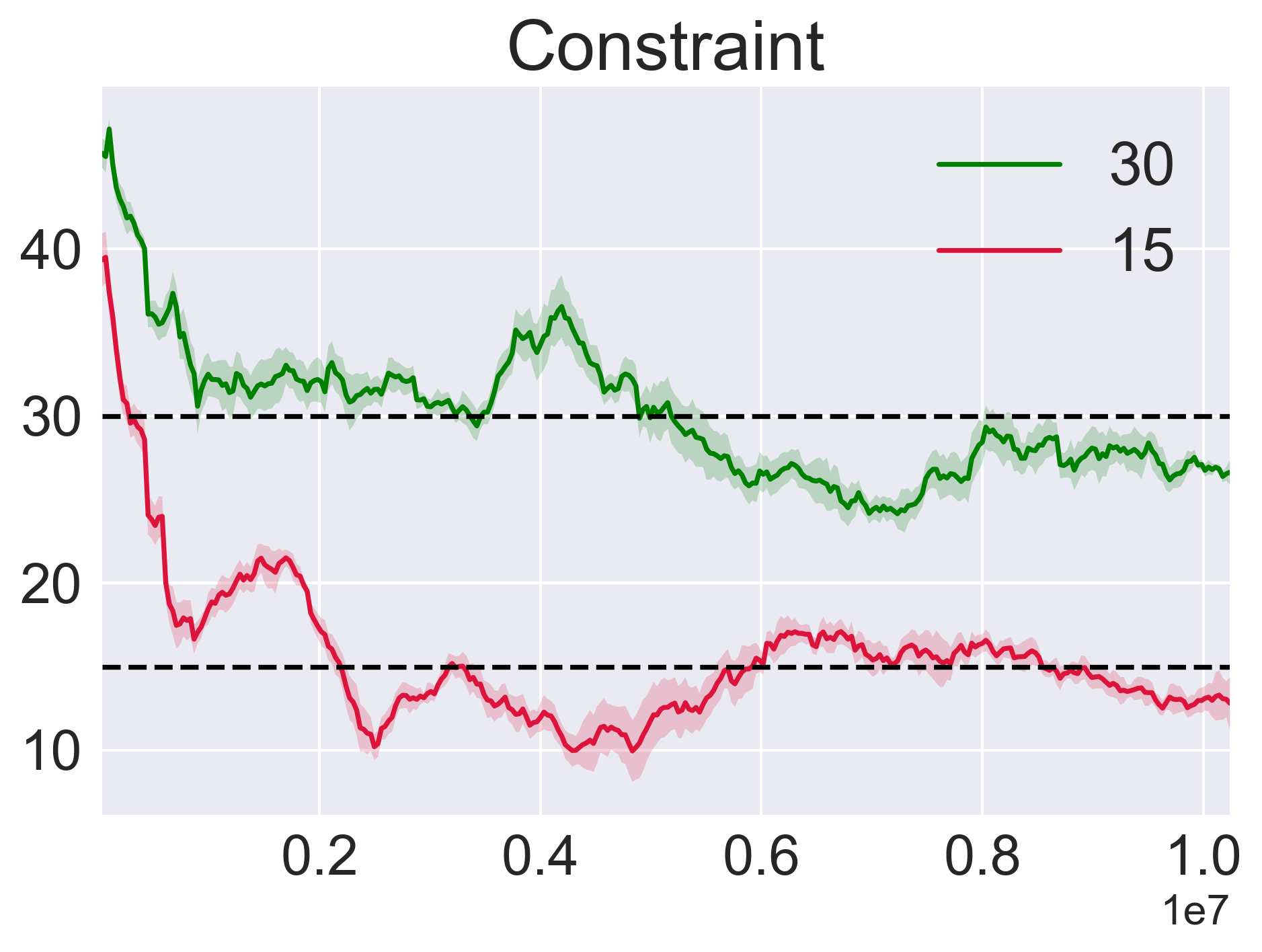}
     \end{subfigure}
    \caption{EPO in PointGoal1-v0 with different cost limits.}
    \label{fig:limit}
\end{figure}

\paragraph{Safety Gymnasium.}
Navigation tasks in Safety Gymnasium with infeasible initial policies are challenging. Figure \ref{fig:gym} presents the learning curves of EPO and baselines. 
Table 1 in the supplemental material shows the mean performance of different algorithms.
EPO demonstrates impressive policy performance and faster constraint satisfaction compared to baselines in challenging tasks. In tasks with infeasible initial policies, EPO converges faster and generates fewer constraint violations, as illustrated in Table 2 in the supplemental material, demonstrating the effectiveness of PMN in providing appropriate penalties when constraints are violated. 
In contrast, IPO tends to generate conservative policies restricted by excessive penalties and struggles with infeasible initial policies for constraint satisfaction.
P3O fails to provide enough penalties when the initial constraint is large, resulting in a slow decrease in constraints.
PPO Lagrangian requires waiting for an increase in the dual variable to mitigate constraints, leading to a slower convergence rate of constraints. 
APPO outperforms PPO Lagrangian in performance and stability but is sensitive to the initial Lagrangian multiplier.
In PointGoal2-v0 and PointButton2-v0, most baselines update the policy conservatively to ensure constraint satisfaction, while EPO shows better performance under the cost limit.

\paragraph{Safety MuJoCo.}
To increase the challenge, we set a cost limit of 25 for all tasks in Safety MuJoCo. Figure \ref{fig:mujoco} illustrates the learning curves of EPO and baseline algorithms.
The results indicate that EPO maintains excellent performance with stable training processes, keeping the policy in the feasible region for tasks with feasible initial policies. 
P3O struggles to provide suitable penalties to keep the policy within the feasible domain, showing poor performance in Swimmer-v4. 
In Safety MuJoCo tasks, the optimal policy often resides near constraint boundaries, while the ReLU operator used in P3O eliminates gradient information around these boundaries, making it challenging to identify the optimal policy. Experimental results demonstrate the effectiveness of our smooth penalty function, leading to enhanced policy performance. 
CPO shows faster performance convergence than EPO in Ant-v4. However, the computational burden imposed by the inverse of the high-dimensional Fisher information matrix leads to training times more than twice as long as EPO.
PPO Lagrangian outperforms EPO in HalfCheetah-v4 and Humanoid-v4 due to the slower updates of Lagrange multipliers compared to policy iterations. The policy is not penalized promptly when it reaches the infeasible region, which leads to higher rewards but large constraint oscillations. 
APPO mitigates these oscillations through the augmented Lagrangian method but decreases policy performance.

\paragraph{Ablation Experiments.}
We conduct ablation experiments to verify the effectiveness of components in EPO. 
First, we compare the PMN with linear and quadratic critics. The results in Figure \ref{fig:ablation}(a)(b) show the EPO with PMN outperforms the other two. The linear critic imposes larger penalties near the constraint boundary, guiding the policy to the feasible region, while the quadratic one imposes larger penalties far from the boundary, accelerating the policy to approach the feasible region. PMN combines the benefits from the linear and quadratic critics, making a trade-off between constraint satisfaction and reward maximization based on the degree of constraint violation.
Second, we compare the smooth EPO function with the non-smooth ReLU function, the results in Figure \ref{fig:ablation}(c)(d) show that the smooth function enhances training stability, especially near constraint boundaries, by providing gradient information across the policy space.

\paragraph{Cost Limit Sensitivity.} We further verify the robustness of EPO at various threshold levels. Figure \ref{fig:limit} indicates that our method can satisfy different constraint limits while maintaining good policy performance.

\section{Conclusion}
In this work, we propose Exterior Penalty Policy Optimization (EPO) for training CRL agents with superior performance and efficient constraint satisfaction.
The Penalty Metric Network (PMN) generates appropriate penalties for fast constraint satisfaction in the infeasible region and safe exploration in the feasible region.
Convergence analysis guarantees consistent constraint satisfaction.
We further provide the worst-case constraint violation and approximation error analysis for the surrogate function, along with a practical implementation of EPO.
EPO is compared with multiple baselines across a wide range of safety tasks, demonstrating its effectiveness in ensuring constraint satisfaction while maintaining policy performance. Future work includes adopting off-policy methods and addressing different types of constraints.

\section*{Acknowledgments}
This work was supported by the National Key Research and Development Plan No. 2022YFB3904204, NSF China under Grant No. 62202299, 62020106005, 61960206002, Shanghai Natural Science Foundation No. 22ZR1429100.

\bibliographystyle{named}
\bibliography{bibfile}

\end{document}